\documentclass{article}
\usepackage[main,final]{neurips_2026}

\usepackage[utf8]{inputenc}
\usepackage[T1]{fontenc}
\usepackage{url}
\usepackage{booktabs}
\usepackage{amsfonts}
\usepackage{nicefrac}
\usepackage{microtype}
\usepackage[table]{xcolor}
\usepackage{graphicx}
\usepackage{caption}
\usepackage{subcaption}
\usepackage{amsmath}
\usepackage{amssymb}
\usepackage{algorithm}
\usepackage{algorithmic}
\usepackage{multirow}
\usepackage{wrapfig}
\usepackage{hyperref}
\hypersetup{
  hidelinks,
  pdftitle={FADE: Fractional Anomalous Dynamics Extrapolation for Training-Free Diffusion Transformers Acceleration},
  pdfauthor={Jinlong Yang, Lizilin, Jinke Wu, Yao Zhou}
}

\title{FADE: Fractional Anomalous Dynamics Extrapolation for Training-Free Diffusion Transformers Acceleration}

\author{%
  Jinlong Yang \\
  Sichuan University \\
  Chengdu, China \\
  \texttt{2023141461083@stu.scu.edu.cn}
  \And
  Lizilin \\
  Sichuan University \\
  Chengdu, China \\
  \texttt{2023141461067@stu.scu.edu.cn}
  \AND
  Jinke Wu \\
  Sichuan University \\
  Chengdu, China \\
  \texttt{w\_rmsl@stu.scu.edu.cn}
  \And
  Yao Zhou\thanks{Corresponding author.} \\
  School of Artificial Intelligence \\
  Sichuan University \\
  Chengdu, China \\
  \texttt{yaozhou@scu.edu.cn}
}

\begin{document}

\maketitle

\begin{abstract}
Diffusion Transformers (DiTs) have established themselves as the preeminent paradigm for high-fidelity generative modeling across various modalities, yet their practical deployment is hindered by the severe computational overhead of iterative sampling. While recent feature-caching methods mitigate this latency through feature reuse or forecasting based on integer-order operators, they remain limited in adapting to the non-stationary and highly curved dynamics of DiT latent trajectories. In this work, we reveal that DiT feature evolution exhibits anomalous diffusion characteristics and a certain degree of spatiotemporal heterogeneity, rendering static, integer-order forecasting operators inadequate and prone to error accumulation. To address this, we propose Fractional Anomalous Dynamics Extrapolation (FADE), a unified framework for training-free diffusion transformers acceleration. FADE models DiT feature evolution via anomalous diffusion and employs a Mittag--Leffler fractional operator to better adapt to high-curvature and non-linear latent trajectories of DiTs. Leveraging the empirical observation that these complex dynamics maintain cross-sample stability, we further construct a Look-Up Table with multiple dimensions through offline profiling. This enables negligible LUT-retrieval overhead, spatiotemporally adaptive parameter retrieval during online inference. Extensive evaluations across diverse architectures and modalities demonstrate that FADE achieves state-of-the-art acceleration, preserving exceptional generation fidelity under aggressive skip intervals with negligible latency overhead.  Our project page is available at \url{https://youngkinlon.github.io/fade-project-page/}.

\end{abstract}
\section{Introduction}
\label{sec:intro}

Diffusion Models~\cite{ho2020denoisingdiffusionprobabilisticmodels} have become a dominant paradigm for generative modeling across images~\cite{chen2023pixartalphafasttrainingdiffusion,cai2025hidreami1highefficientimagegenerative,flux2024}, videos~\cite{kong2024hunyuanvideo,wan2025}, and audio~\cite{kong2021diffwaveversatilediffusionmodel,liu2023audioldm}, owing to their strong high-fidelity synthesis capability. Recent advances in Diffusion Transformers (DiTs)~\cite{peebles2023scalablediffusionmodelstransformers} have further improved this capability, but also introduce substantial computational overhead. Their iterative sampling process requires sequential forward passes through large-scale models, resulting in high latency that hinders practical deployment and motivates training-free acceleration without modifying model weights.

To alleviate this bottleneck, feature caching has emerged as a training-free acceleration strategy. Traditional caching methods~\cite{ma2023deepcacheacceleratingdiffusionmodels,selvaraju2024FORAfastforwardcachingdiffusion} reuse features from previous timesteps by exploiting temporal consistency in intermediate representations. However, this cache-and-reuse paradigm degrades generation quality under large skip intervals, as feature similarity diminishes over longer temporal distances~\cite{selvaraju2024FORAfastforwardcachingdiffusion, ma2023deepcacheacceleratingdiffusionmodels}. Methods such as TaylorSeer~\cite{liu2025reusingforecastingacceleratingdiffusion} and HiCache~\cite{feng2026hicachepluginscaledhermiteupgrade} formulate feature forecasting based on integer-order forecasting operators. Yet such formulations implicitly assume locally smooth dynamics with a fixed coefficient law, providing only limited approximations to DiT latent trajectories with complex geometry. As a result, they struggle to capture the intrinsic curvature, stage-dependent variations, and abrupt trajectory changes observed across denoising stages.
\begin{figure}[t]
  \centering
  \begin{minipage}{0.8\linewidth}
    \centering
    \begin{subfigure}[t]{0.46\linewidth}
      \centering
      \includegraphics[width=\linewidth]{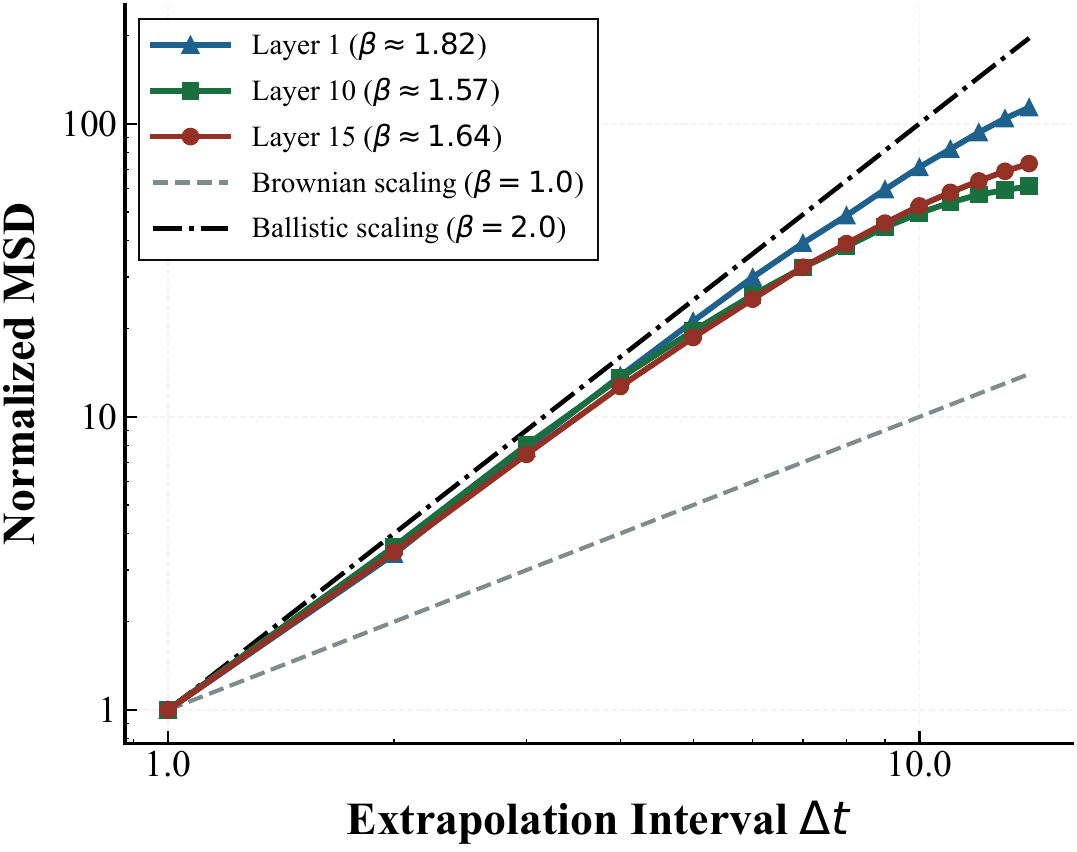}
      \caption{Anomalous diffusion scaling}
      \label{fig:msd_scaling}
    \end{subfigure}
    \hfill
    \begin{subfigure}[t]{0.46\linewidth}
      \centering
      \includegraphics[width=\linewidth]{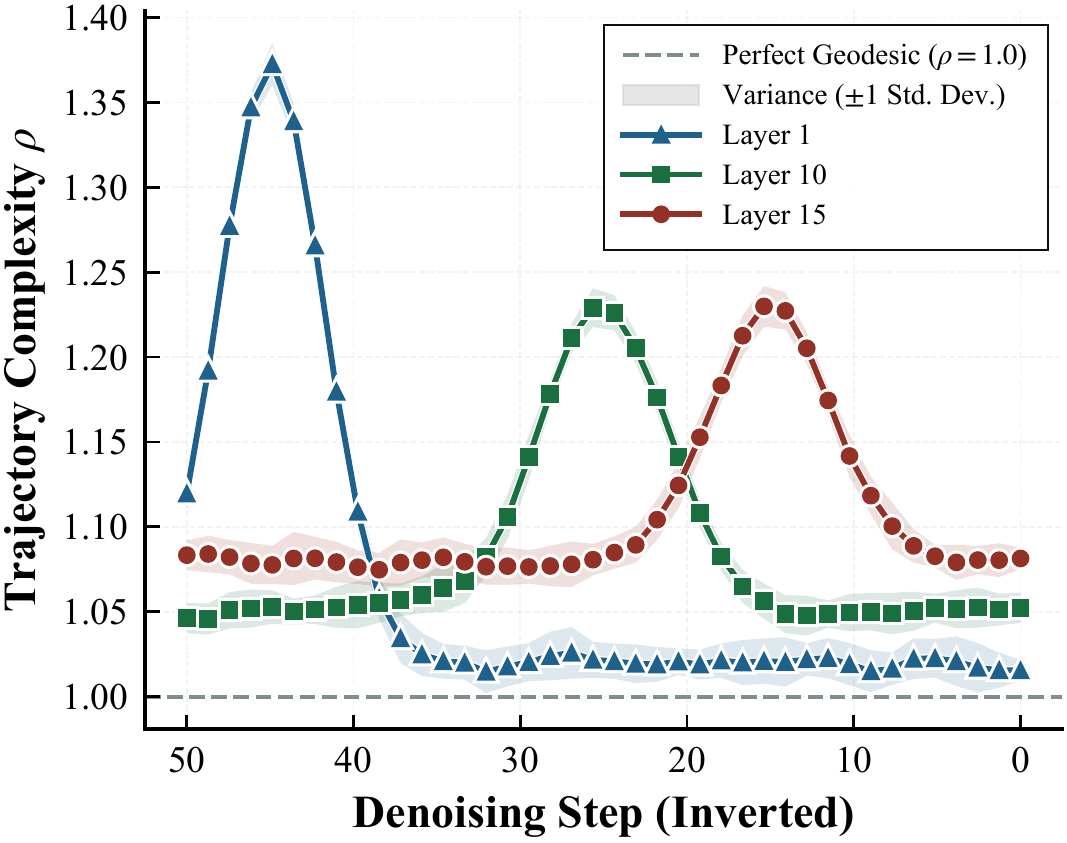} 
      \caption{Spatiotemporal trajectory consistency} 
      \label{fig:phase_transition} 
    \end{subfigure} 
  \end{minipage}  
   \caption{Empirical evidence of anomalous and structured dynamics in DiT. 
    (a) The normalized mean squared displacement across layers follows an anomalous diffusion scaling law.
    (b) Cross-sample trajectory consistency across different layers and denoising steps.} 
  \label{fig:empirical_dynamics} 
\end{figure}

This limitation reflects a deeper geometric mismatch between existing forecasting schemes and DiT latent dynamics. Prior studies have shown that flat Euclidean assumptions are insufficient to capture the topology of deep generative latent spaces~\cite{kalatzis2020variational}, while recent analyses further suggest the existence of fractal phase-transition structures in diffusion latents~\cite{lobashev2025hessian}. Since anomalous diffusion commonly arises on fractal structures~\cite{oliveira2019anomalous}, this connection motivates us to investigate whether DiT feature evolution exhibits anomalous diffusion behavior.
In anomalous diffusion studies, the power-law scaling exponent of the mean squared displacement (MSD) serves as a standard empirical diagnostic for deviations from Brownian diffusion~\cite{METZLER20001}.
Using this diagnostic, we observe that the normalized MSD of DiT features deviates from Brownian linear scaling, as shown in Fig.~\ref{fig:empirical_dynamics}(a), thereby supporting a fractional perspective for modeling such dynamics. To further characterize the structure of feature evolution, we examine its variation across denoising stages, network layers, and input samples from an anomalous-diffusion perspective. As shown in Fig.~\ref{fig:empirical_dynamics}(b), feature dynamics vary substantially across layers and timesteps, while exhibiting relatively smaller variation across input samples. This observation suggests structured stage- and layer-dependent dynamics with a degree of cross-sample consistency, providing empirical support for calibrating forecasting parameters offline by denoising stage and network location and reusing them across samples. 

Building on these observations, our key insight is to treat DiT feature evolution as an anomalous diffusion process over complex latent geometry, while exploiting its empirical cross-sample regularity for offline calibration. This motivates a forecasting framework that combines a fractional dynamical model with offline-calibrated stage-aware adaptation. Accordingly, we propose FADE (Fractional Anomalous Dynamics Extrapolation), a training-free framework that models DiT feature evolution via anomalous diffusion for diffusion transformers acceleration. A Mittag--Leffler fractional operator is introduced to generalize the classical integer-order shift operator, enabling flexible prediction of non-Brownian feature evolution. To operationalize this adaptive formulation, FADE profiles the optimal fractional parameters offline and stores them in a multi-dimensional LUT, which allows stage-aware feature forecasting through $\mathcal{O}(1)$ retrieval during inference. This design improves extrapolation robustness under large skip intervals while preserving generation quality with negligible overhead.

In summary, the principal contributions of our work are as follows:
\begin{itemize}
     \item \textbf{Fractional Anomalous Dynamics Extrapolation (FADE):} 
    We propose a training-free forecasting framework that models DiT feature evolution as an anomalous diffusion process. By generalizing the classical exponential shift operator with a Mittag--Leffler fractional operator, FADE better adapts to the complex, high-curvature dynamics of latent trajectories.
    \item \textbf{Zero-Overhead Adaptive Forecasting:} To translate our fractional framework into practical acceleration, we construct a Look-Up Table indexed by multiple dimensions through offline profiling. This design enables spatiotemporally adaptive parameter retrieval in constant time, ensuring robust feature forecasting without introducing any additional inference latency.
    \item \textbf{Empirical Characterization of DiT Feature Dynamics:} We reveal that DiT feature evolution exhibits anomalous diffusion behavior and structured spatiotemporal heterogeneity across denoising stages and network locations. Meanwhile, these dynamics remain relatively stable across input samples, providing empirical support for offline calibration.
    \item \textbf{State-of-the-Art Acceleration:} Extensive evaluations across diverse architectures and modalities, including DiT-XL/2, FLUX.1, and HunyuanVideo, demonstrate that FADE consistently outperforms existing baselines, maintaining exceptional generation fidelity and negligible latency overhead even under long skip intervals.
\end{itemize}
\section{Related Work}
\label{sec:formatting}

\paragraph{Computation Reduction Strategies.}
Previous efforts to accelerate diffusion models focus on either reducing the number of sampling steps or compressing the denoising network. Step-reduction methods accelerate the generative process through efficient ODE solvers like DDIM~\cite{song2022denoisingdiffusionimplicitmodels} and DPM-Solver~\cite{lu2022dpmsolverfastodesolver,zhao2023unipcunifiedpredictorcorrectorframework,Lu_2025}, knowledge distillation~\cite{salimans2022progressivedistillationfastsampling,luo2023latentconsistencymodelssynthesizing}, or few-step paradigms such as Rectified Flow~\cite{liu2022flowstraightfastlearning} and Consistency Models~\cite{song2023consistencymodels}. However, these approaches often degrade generation quality or require expensive retraining~\cite{ma2024sitexploringflowdiffusionbased}. Alternatively, network compression techniques such as pruning~\cite{fang2023structuralpruningdiffusionmodels,yuan2024ditfastattn}, quantization~\cite{li2023qdiffusionquantizingdiffusionmodels,shang2023ptqdm,yang2025lrq}, and token merging~\cite{zhang2024tokenpruningcachingbetter,bolya2023tokenmergingfaststable} reduce the FLOPs of each denoising step. However, their efficiency gains often come at the cost of reduced model expressivity, which may compromise generation fidelity under aggressive compression. In contrast, feature caching reduces redundant intermediate feature computation to accelerate inference.

\paragraph{Feature Caching.}
As a prominent training-free alternative, feature caching skips redundant intermediate computations. Existing methods generally fall into two paradigms: cache-then-reuse and cache-then-forecast. Reuse-based approaches reuse cached representations, including direct reuse methods such as DeepCache~\cite{ma2023deepcacheacceleratingdiffusionmodels}, PAB~\cite{zhao2024real}, and FORA~\cite{selvaraju2024FORAfastforwardcachingdiffusion}, token-level methods such as ToCa~\cite{zou2025acceleratingdiffusiontransformerstokenwise}, Tokencache~\cite{lou2024token}, and DaTo~\cite{zhang2024token}, and error-aware or adaptive caching variants such as TeaCache~\cite{liu2025timestep}, AdaCache~\cite{kahatapitiya2025adaptive}, and ERTACache~\cite{peng2025ertacache}. MEDA~\cite{11711089} uses multiobjective evolutionary search
to optimize feature-reuse schedules, balancing computational
cost and generation quality.
However, larger skip intervals inevitably enlarge the temporal
representation mismatch, degrading generation quality. However, larger skip intervals inevitably enlarge the temporal representation mismatch, degrading generation quality. Forecasting-based approaches, pioneered by TaylorSeer~\cite{liu2025reusingforecastingacceleratingdiffusion}, actively predict feature evolution. Recent advancements seek to refine this paradigm: HiCache~\cite{feng2026hicachepluginscaledhermiteupgrade} improves the standard Taylor expansion by introducing scaled Hermite coefficients, while FOCA~\cite{zheng2025forecastcalibratefeaturecaching} drives the forecasting process by utilizing both the features themselves and their first-order finite-difference derivatives. BRACE~\cite{yang2026bracetamingsharpirregularities} further explores barycentric rational forecasting with adapted Chebyshev weights, directly aggregating cached historical features to accommodate local trajectory irregularities. Despite these algorithmic refinements, these methods treat the non-equilibrium denoising trajectory as a globally uniform dynamical system, limiting their capacity to characterize the anomalous diffusion properties and complex temporal phase transitions of latent features. This ultimately results in trajectory deviation under large skip intervals.

\section{Method}

\subsection{Preliminaries}
Diffusion models generate data by iteratively reversing a predefined forward noising process via a neural network $\epsilon_\theta(\mathbf{x}_t, t)$ that estimates the denoising direction at each timestep. Starting from Gaussian noise, the latent variable $\mathbf{x}_t$ is progressively refined over $T$ timesteps toward a clean sample:
\begin{equation}
    \mathbf{x}_{t-1} = \frac{1}{\sqrt{\alpha_t}} \left( \mathbf{x}_t - \frac{1 - \alpha_t}{\sqrt{1 - \bar{\alpha}_t}} \epsilon_\theta(\mathbf{x}_t, t) \right) + 
    \sigma_t \boldsymbol{\epsilon}.
    \label{eq:reverse_process}
\end{equation}

Since diffusion sampling requires $T$ sequential model evaluations, inference is costly, especially for large-scale DiTs~\cite{peebles2023scalablediffusionmodelstransformers}. To reduce full forward passes at selected timesteps, \textit{Predictive feature caching} reduces this cost by forecasting intermediate features $\mathbf{F}^l$ at layer $l$. With a recursive discrete difference operator $\Delta$, the $m$-th order Taylor predictor over distance $\delta$ is:
\begin{equation}
\mathbf{F}_{\text{pred}}(\mathbf{x}^l_{t-\delta}) = \mathbf{F}(\mathbf{x}^l_t) + \sum_{i=1}^{m} \frac{\Delta^i \mathbf{F}(\mathbf{x}^l_t)}{i!} (-\delta)^i.
\label{eq:taylor}
\end{equation}
Based on operator calculus, the above polynomial extrapolation over a prediction distance $\delta$ can be interpreted as applying the exponential shift operator $e^z$ to the current feature $\mathbf{F}_t$, where $z = -\delta \nabla_t$ and $\nabla_t = \partial / \partial t$ denotes the temporal derivative operator:
\begin{equation}
    \mathbf{F}_{t-\delta} \approx e^z \mathbf{F}_t.
    \label{eq:operator_taylor}
\end{equation}
From this perspective, conventional integer-order extrapolation methods typically rely on finite-order local approximations of the shift operator, which may struggle to capture the complex and non-stationary dynamics of DiT's latent trajectories.

\subsection{Overall Framework of FADE}
\label{subsec:framework}

\begin{figure*}[t]
    \centering
    \includegraphics[width=0.95\textwidth]{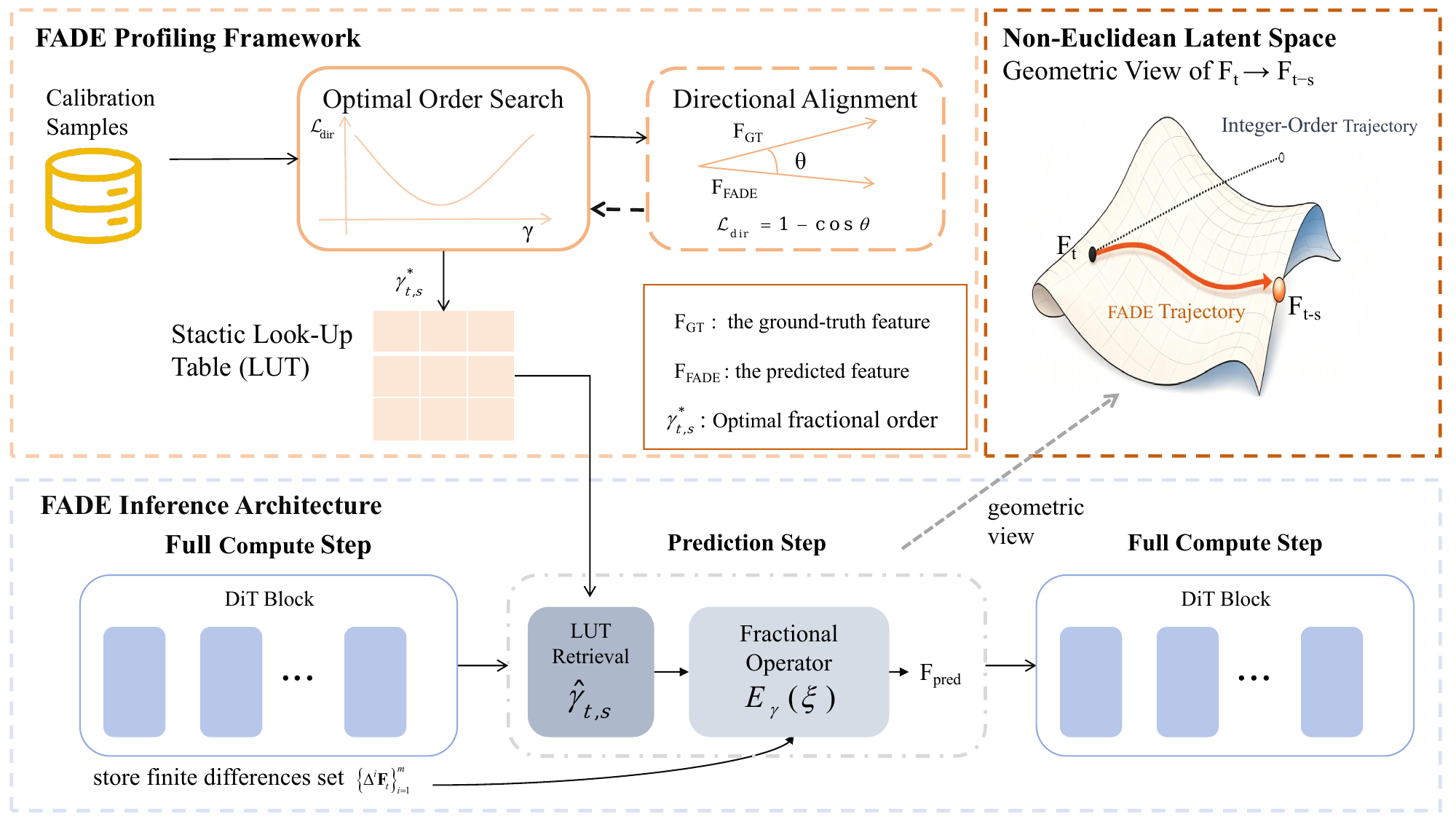}
    \caption{\textbf{Overall framework of FADE.} FADE consists of an offline profiling stage and an online inference stage. 
The offline stage calibrates and stores the optimal fractional orders in a static LUT, while the online stage alternates between full-compute and prediction steps using LUT-retrieved orders for feature forecasting. 
The upper-right panel illustrates the geometric contrast between the FADE trajectory and the integer-order trajectory in the latent space.}
    \label{fig:fade_framework}
\end{figure*}

Motivated by the anomalous diffusion of DiT latent feature evolution, we propose Fractional Anomalous Dynamics Extrapolation (FADE). As shown in Figure~\ref{fig:fade_framework}, FADE consists of an offline \emph{FADE Profiling Framework} and an online \emph{FADE Inference Architecture}. In the profiling stage, calibration samples are used for \emph{Optimal Order Search} with \emph{Directional Alignment}, where the optimal fractional order $\gamma^{*}_{t,s}$ is obtained by minimizing the directional loss $\mathcal{L}_{\mathrm{dir}}$ between the predicted feature $\mathbf{F}_{\mathrm{FADE}}$ and the ground-truth feature $\mathbf{F}_{\mathrm{GT}}$. The calibrated orders are then stored in a static look-up table (LUT), where $s$ indexes the network location(e.g., layer, module, stream, etc). In the inference stage, denoising follows a cache interval $N$ and alternates between \emph{Full Compute Steps} and \emph{Prediction Steps}: Full Compute Steps perform forward passes and cache the finite-difference set $\{\Delta^{i}\mathbf{F}_{t}\}_{i=1}^{m}$, while Prediction Steps retrieve $\hat{\gamma}_{t,s}$ from the LUT and apply the \emph{Fractional Operator} $E_{\gamma}(\xi)$ to generate the predicted feature $\mathbf{F}_{\mathrm{pred}}$. The geometric view in Figure~\ref{fig:fade_framework} further illustrates that FADE better adapts to the complex latent manifold of DiTs. 

\subsection{Modeling DiT Feature Evolution via Anomalous Diffusion}
\label{subsec:derivation}
Building upon the observed latent dynamics and prior studies on
DiT latent geometry, we model DiT feature evolution from an
anomalous-diffusion perspective. Anomalous transport on fractal
structures provides a geometric motivation for this view~\cite{oliveira2019anomalous},
while our observed deviations from Brownian MSD scaling provide
empirical support. To formulate a forecasting rule for such
dynamics, we draw on classical fractional dynamics, where
Mittag--Leffler functions commonly arise in propagation and
relaxation solutions~\cite{METZLER20001}. Since the Mittag--Leffler
function reduces to the exponential function at unit order,
we adopt it as a fractional operator prior to generalize the
classical exponential shift operator. This provides a flexible
coefficient law for controlling how higher-order local variations
contribute to feature prediction:

\begin{equation}
E_\gamma(\xi) = \sum_{k=0}^{\infty} \frac{\xi^k}{\Gamma(\gamma k + 1)}.
\label{eq:mittag_leffler}
\end{equation}
Here, $\Gamma(\cdot)$ denotes the Gamma function used to parameterize the fractional coefficients. Following this fractional coefficient law, we construct a finite-order feature predictor as
\begin{equation}
\mathbf{F}_{\text{pred}}(\mathbf{x}^l_{t-\delta}) 
= \mathbf{F}(\mathbf{x}^l_t) 
+ \sum_{i=1}^{m} 
\frac{\Delta^i \mathbf{F}(\mathbf{x}^l_t)}{\Gamma(\gamma i + 1)} (-\delta)^i.
\label{eq:fade_extrapolation}
\end{equation}

Under this formulation, the fractional order $\gamma$ controls the coefficient law used for feature extrapolation. 
Since DiT denoising is highly non-stationary, feature dynamics vary across timesteps and latent states, often shifting between sub- and super-diffusive-like behaviors. 
This makes a fixed integer-order rule insufficient for modeling local evolution. 
FADE therefore assigns state-specific fractional orders to adapt the extrapolation law to heterogeneous anomalous dynamics. 
To obtain these orders without runtime search, we introduce a Stage-Aware Calibration Strategy.

\subsection{Stage-Aware Calibration Strategy}
As established in our empirical analysis, existing static caching frameworks overlook a key structural property of Diffusion Transformers: feature dynamics vary substantially across latent states and timesteps, yet exhibit a certain degree of cross-sample regularity. Leveraging this cross-sample consistency, we propose a Stage-Aware Calibration Strategy that offline pre-computes the optimal fractional order $\gamma^*_{t,s}$ for each state $s$ and timestep $t$, thereby avoiding online search overhead with a short one-time calibration stage and negligible LUT storage. 

\paragraph{Directional Monitoring Objective.}
Calibration is formulated as a local trajectory-matching problem.
We use cosine distance to emphasize directional agreement
between predicted and reference features. A purely directional objective is therefore adopted, and $\mathcal{L}_{\text{dir}}$ is defined as the cosine distance between the predicted feature $\mathbf{F}_{\text{FADE}}(\gamma)$ and the ground-truth feature $\mathbf{F}_{\text{GT}}$:
\begin{equation}
    \mathcal{L}_{\text{dir}}(\gamma) = 1 - \frac{\mathbf{F}_{\text{FADE}}(\gamma) \cdot \mathbf{F}_{\text{GT}}}{\|\mathbf{F}_{\text{FADE}}(\gamma)\|_2 \|\mathbf{F}_{\text{GT}}\|_2}.
    \label{eq:directional_loss}
\end{equation}

\paragraph{Optimal Order Search and Global Refinement.}
Using a small calibration dataset $\mathcal{D}_{\text{calib}}$, the optimal fractional order is obtained by minimizing the expected directional error over a discrete search grid $\mathcal{G}$, which spans both sub- and super-diffusion-like regimes:
\begin{equation}
    \gamma^*_{t,s} = \underset{\gamma \in \mathcal{G}}{\arg\min} \ \mathbb{E}_{\mathbf{x} \sim \mathcal{D}_{\text{calib}}} \left[ \mathcal{L}_{\text{dir}}\left(\mathbf{F}_{\text{FADE}}(\gamma); \mathbf{x}, t, s\right) \right].
    \label{eq:gamma_optimization}
\end{equation}

The offline-calibrated fractional orders $\gamma^*_{t,s}$ may
still exhibit a global scale bias in practical inference.
To account for this potential bias, we introduce a global
refinement factor $\lambda$ and use
$\hat{\gamma}_{t,s} = \lambda \cdot \gamma^*_{t,s}$,
providing a lightweight global adjustment to the calibrated orders. The refined orders $\hat{\gamma}_{t,s}$ are then stored in the LUT described in Sec.~\ref{subsec:framework} for inference-time retrieval. Detailed calibration settings are provided in Appendix~\ref{sec:calibration_details}.

\subsection{Error Analysis from a Fractional Perspective}
\label{subsec:error_analysis_main}

We provide a local and conditional error analysis under a
fractional model of local latent evolution.
Assume that the local ground-truth evolution admits the
following representation:

\begin{equation}
\mathbf{F}^{\mathrm{GT}}_{t-\delta,s}
=
\mathbf{F}_{t,s}
+
\sum_{i=1}^{m}
\frac{\Delta^i \mathbf{F}_{t,s}}{\Gamma(\gamma^\dagger_{t,s} i + 1)}
(-\delta_{\mathcal{M},t,s})^i
+
\mathbf{R}^{(t,s)}_{m+1},
\qquad
\|\mathbf{R}^{(t,s)}_{m+1}\|
\le
C_{t,s}\,\delta_{\mathcal{M},t,s}^{\,m+1},
\label{eq:error_assumption_main}
\end{equation}
where $\delta_{\mathcal{M},t,s}$ denotes the intrinsic propagation
scale along the local feature trajectory, which reflects the
underlying distance traveled by features on the latent manifold,
while $\delta$ denotes the prediction distance used in practice.
This distinction captures a potential propagation-scale mismatch.
The profiled order $\gamma^*_{t,s}$ is used as an estimate of the
intrinsic order $\gamma^\dagger_{t,s}$, and FADE applies the globally
refined order $\hat{\gamma}_{t,s}=\lambda\gamma^*_{t,s}$ at inference
time. Substituting Eq.~\eqref{eq:fade_extrapolation} into
Eq.~\eqref{eq:error_assumption_main}, we obtain
\begin{equation}
\|\mathbf{F}^{\mathrm{GT}}_{t-\delta,s}-\widehat{\mathbf{F}}^{\mathrm{FADE}}_{t-\delta,s}\|
\le
C_{t,s}\,\delta_{\mathcal{M},t,s}^{\,m+1}
+
\sum_{i=1}^{m}
\left\|
\Delta^i \mathbf{F}_{t,s}
\right\|
\left|
\frac{(-\delta_{\mathcal{M},t,s})^i}{\Gamma(\gamma^\dagger_{t,s} i+1)}
-
\frac{(-\delta)^i}{\Gamma(\lambda\gamma^*_{t,s} i+1)}
\right|.
\label{eq:fade_error_bound_main}
\end{equation}
The second term combines the intrinsic--extrinsic scale mismatch and the coefficient mismatch. For Taylor extrapolation in Eq.~\eqref{eq:taylor}, the corresponding mismatch is
\begin{equation}
\sum_{i=1}^{m}
\left\|
\Delta^i \mathbf{F}_{t,s}
\right\|
\left|
\frac{(-\delta_{\mathcal{M},t,s})^i}{\Gamma(\gamma^\dagger_{t,s} i+1)}
-
\frac{(-\delta)^i}{i!}
\right|.
\label{eq:taylor_error_bound_main}
\end{equation}
Since Taylor extrapolation corresponds to the $\gamma=1$ case of the fractional coefficient family, it can be viewed as a special case of FADE. In contrast, FADE calibrates a stage-adaptive order $\gamma^*_{t,s}$ to accommodate the heterogeneous dynamics across DiT timesteps and network states, and applies a refined order $\hat{\gamma}_{t,s}=\lambda\gamma^*_{t,s}$ at inference time. The global factor $\lambda$ can be viewed as a lightweight correction
that helps accommodate residual propagation-scale mismatch.
Together, these designs provide a more flexible local approximation
of the intrinsic dynamics, consistent with the improved robustness
observed under large prediction distances. 
\section{Experiments}

\subsection{Experimental Setup}

FADE is evaluated across three representative generative tasks:
(i) class-conditional image generation on
ImageNet-1K~\cite{deng2009imagenet} using
DiT-XL/2~\cite{peebles2023scalablediffusionmodelstransformers};
(ii) text-to-image generation on
DrawBench~\cite{saharia2022photorealistictexttoimagediffusionmodels}
using the rectified flow-based
FLUX.1-dev~\cite{labs2025flux1kontextflowmatching,flux2024,liu2022flowstraightfastlearning};
and (iii) text-to-video generation using
HunyuanVideo~\cite{kong2024hunyuanvideo} on the HunyuanLarge
architecture. For DiT-XL/2, we generate images at
$256\times256$ resolution using the DDIM sampler with
50 denoising steps. For FLUX.1-dev, we use the Flow Match
Euler scheduler with 50 sampling steps at
$1024\times1024$ resolution. HunyuanVideo follows its
official inference pipeline with 50 flow-matching sampling
steps and default video generation settings.

We compare FADE with representative training-free caching
methods, including
FORA~\cite{selvaraju2024FORAfastforwardcachingdiffusion},
TaylorSeer~\cite{liu2025reusingforecastingacceleratingdiffusion},
and HiCache~\cite{feng2026hicachepluginscaledhermiteupgrade}.
Within each comparison, caching methods use the same
pretrained checkpoint, sampler, number of sampling steps,
resolution, guidance settings, and evaluation inputs.
Qualitative comparisons additionally use matched random
seeds and acceleration schedules to facilitate direct
visual assessment.

Quantitative performance is assessed using
FID-50K~\cite{fid}, sFID, and Inception Score for DiT-XL/2.
For FLUX.1-dev, we report
ImageReward~\cite{xu2023imagereward},
CLIP Score~\cite{hessel2022clipscorereferencefreeevaluationmetric},
and CycleReward~\cite{Bahng_2025_ICCV}, together with
PSNR, SSIM~\cite{PSNRSSIM}, and
LPIPS~\cite{LPIPsZhang_2018_CVPR} to assess consistency
with the full-compute baseline.
HunyuanVideo is evaluated using the 16-dimensional
VBench suite~\cite{huang2023vbenchcomprehensivebenchmarksuite}.
Computational efficiency is assessed using FLOPs and
measured generation latency. FLOPs reduction and wall-clock
speedup are computed separately as the ratios of baseline
FLOPs and latency to those of the accelerated method,
respectively. Offline calibration cost is reported
separately from inference latency.

\subsection{Empirical Validation of Fractional Dynamics}
\label{sec:fractional_dynamics}

\begin{figure}[t]
  \centering
  \begin{minipage}{\linewidth} 
    \begin{subfigure}[t]{0.48\linewidth}
      \centering
      \includegraphics[width=\linewidth]{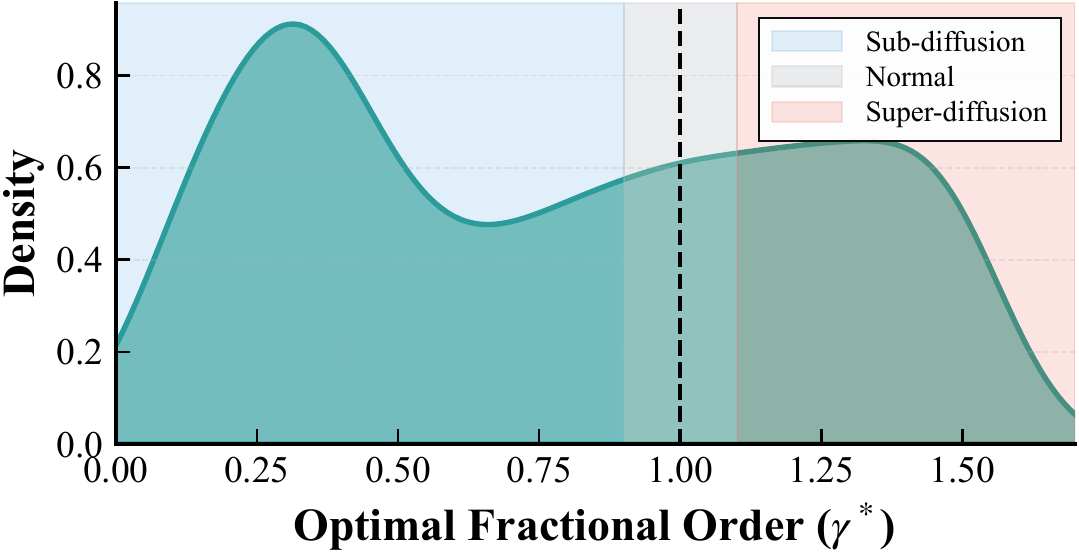}
      \caption{Global distribution of $\gamma^*$.}
    \end{subfigure}
    \hfill
    \begin{subfigure}[t]{0.48\linewidth}
      \centering
      \includegraphics[width=\linewidth]{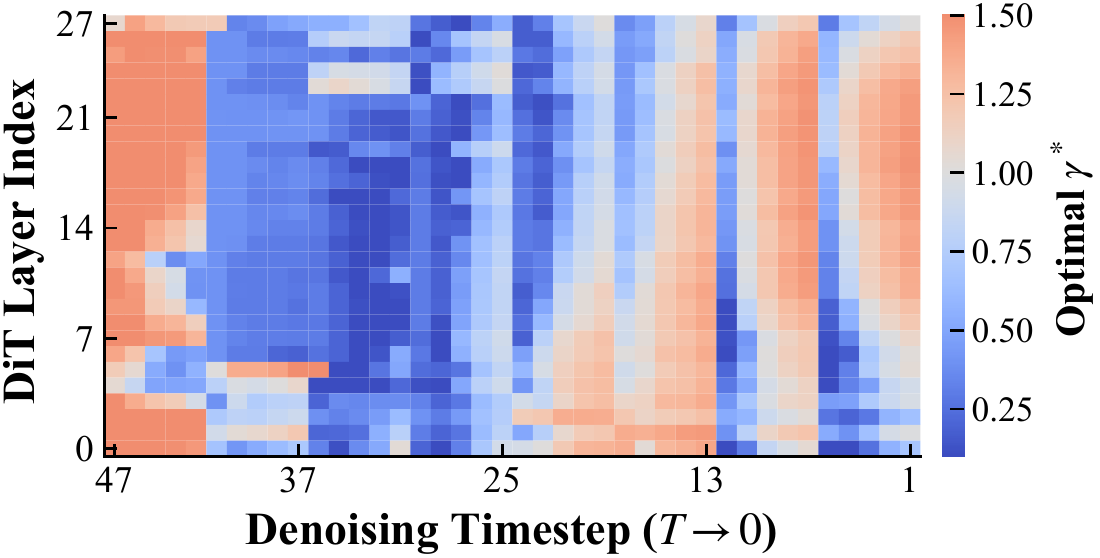}
      \caption{Spatiotemporal heatmap of $\gamma^*$.}
    \end{subfigure}
  \end{minipage}
  \caption{\textbf{Empirical validation of fractional dynamics on DiT-XL/2.}
  The optimal fractional orders deviate markedly from the static integer-order assumption and exhibit a pronounced non-stationary temporal pattern across denoising timesteps.}
  \label{fig:dit_fractional_dynamics}
\end{figure}

To validate the fractional prior underlying FADE, we profile the optimal fractional orders $\gamma^*$ across representative architectures and modalities. If DiT feature evolution were well described by conventional integer-order dynamics, the optimal states would concentrate around $\gamma=1$, which is not observed in practice. On DiT-XL/2, Fig.~\ref{fig:dit_fractional_dynamics} shows that the global distribution of $\gamma^*$ is clearly shifted away from $\gamma=1$, indicating that standard integer-order extrapolation captures only a limited portion of the effective generative states. The spatiotemporal heatmap further reveals a U-shaped temporal pattern: the middle denoising phase is more sub-diffusion-like, whereas the early and late phases are more super-diffusion-like in a phenomenological sense.

This heterogeneity also extends across architectures. FLUX.1 shows an inverted temporal trend and clear DoubleStream--SingleStream decoupling, while HunyuanVideo exhibits a broader super-diffusion-like tendency. Together, these results suggest that optimal feature dynamics are non-integer, architecture-dependent, and stage-dependent, motivating the stage-aware calibration design of FADE. More details are provided in Appendix~\ref{sec:dynamic_analysis}. Meanwhile, the cross-sample variance of $\gamma^*$ is consistently small on DiT-XL/2 and remains within an acceptable range on FLUX.1 and HunyuanVideo.

\subsection{Class-Conditional Image Generation}
\label{sec:exp_discussion}
\begin{table}[t]
    \centering
    \caption{Quantitative comparison of class-conditional image generation with DiT-XL/2}
    \label{tab:main_results}
    \small                 
    \setlength{\tabcolsep}{1.5pt}  
    \resizebox{0.8\columnwidth}{!}{   
    \begin{tabular}{l c c c c c c}
        \toprule
        \textbf{Method}
        & \textbf{Lat. (s)$\downarrow$}
        & \textbf{FLOPs (T)$\downarrow$}
        & \textbf{Spd.$\uparrow$}
        & \textbf{FID$\downarrow$}
        & \textbf{sFID$\downarrow$}
        & \textbf{IS$\uparrow$} \\
        \midrule
        DDIM-50 steps~\cite{song2022denoisingdiffusionimplicitmodels} & 1.18 & 23.74 & 1.00$\times$ & 2.29 & 4.29 & 237.31 \\
        DDIM-25 steps~\cite{song2022denoisingdiffusionimplicitmodels} & 0.62 & 11.87 & 2.00$\times$ & 2.91 & 4.57 & 232.22 \\
        \midrule
        DDIM-15 steps~\cite{song2022denoisingdiffusionimplicitmodels} & 0.42 & 7.12 & 3.33$\times$ & 5.09 & 6.11 & 205.54 \\
        FORA($N$=4)~\cite{selvaraju2024FORAfastforwardcachingdiffusion} & 0.45 & 6.66 & 3.56$\times$ & 4.73 & 8.45 & 214.20 \\
        Taylor ($N$=4, $m$=4)~\cite{liu2025reusingforecastingacceleratingdiffusion} & 0.61 & 6.66 & 3.56$\times$ & 2.59 & 5.18 & 232.90 \\
        HiCache ($N$=4, $m$=4)~\cite{feng2026hicachepluginscaledhermiteupgrade} & 0.57 & 6.66 & 3.56$\times$ & 2.51 & 5.20 & 232.72 \\
        \rowcolor{gray!15}
        \textbf{FADE($N$=4, $m$=4)} & 0.55 & \textbf{6.66} & \textbf{3.56}$\times$ & \textbf{2.41} & \textbf{4.79} & \textbf{236.76} \\
        \midrule
        DDIM-12 steps~\cite{song2022denoisingdiffusionimplicitmodels} & 0.33 & 5.70 & 4.17$\times$ & 7.94 & 8.07 & 183.88 \\
        FORA($N$=5)~\cite{selvaraju2024FORAfastforwardcachingdiffusion} & 0.37 & 5.24 & 4.53$\times$ & 5.81 & 9.61 & 199.96 \\
        Taylor ($N$=5, $m$=4)~\cite{liu2025reusingforecastingacceleratingdiffusion} & 0.55 & 5.24 & 4.53$\times$ & 2.73 & 5.34 & 228.75 \\
        HiCache ($N$=5, $m$=4)~\cite{feng2026hicachepluginscaledhermiteupgrade} & 0.50 & 5.24 & 4.53$\times$ & 2.67 & 5.48 & 229.44 \\
        \rowcolor{gray!15}
        \textbf{FADE($N$=5, $m$=4)} & 0.49 & \textbf{5.24} & \textbf{4.53}$\times$ & \textbf{2.55} & \textbf{5.15} & \textbf{230.89} \\
        \midrule
        DDIM-10 steps~\cite{song2022denoisingdiffusionimplicitmodels} & 0.30 & 4.75 & 5.00$\times$ & 12.17 & 11.24 & 159.20 \\
        FORA($N$=6)~\cite{selvaraju2024FORAfastforwardcachingdiffusion} & 0.34 & 4.76 & 4.98$\times$ & 9.22 & 14.86 & 169.02 \\
        Taylor ($N$=6, $m$=4)~\cite{liu2025reusingforecastingacceleratingdiffusion} & 0.50 & 4.76 & 4.98$\times$ & 3.17 & 6.36 & 222.00  \\
        HiCache ($N$=6, $m$=4)~\cite{feng2026hicachepluginscaledhermiteupgrade} & 0.46 & 4.76 & 4.98$\times$ & 3.09 & 6.30 & 220.55 \\
        \rowcolor{gray!15}
        \textbf{FADE($N$=6, $m$=4)} & 0.45 & \textbf{4.76} & \textbf{4.98}$\times$ & \textbf{2.87} & \textbf{5.73} & \textbf{225.78} \\
        \bottomrule
    \end{tabular}
    }
    
\end{table}

As demonstrated in Table~\ref{tab:main_results}, FADE consistently dominates both generation quality and efficiency metrics across all acceleration ratios on DiT-XL/2. When operating at a caching interval of $N=4$, FADE achieves an outstanding FID of 2.41 and an sFID of 4.79. This not only crushes the 15-step DDIM (FID 5.09) with fewer FLOPs, but also outperforms TaylorSeer's~\cite{liu2025reusingforecastingacceleratingdiffusion} FID of 2.59 and HiCache's~\cite{feng2026hicachepluginscaledhermiteupgrade} 2.51 under strictly identical computational costs. This superiority becomes even more pronounced under aggressive acceleration regimes. When scaling to $N=6$, FADE maintains a highly competitive FID of 2.87, completely mitigating the severe structural collapse observed in FORA~\cite{selvaraju2024FORAfastforwardcachingdiffusion} and standard Taylor caching~\cite{liu2025reusingforecastingacceleratingdiffusion}, which suffer from severely degraded FIDs of 9.22 and 3.17 respectively.

\subsection{Text-to-Image Generation}
\begin{figure}[t]
    \centering
    \includegraphics[width=\textwidth]{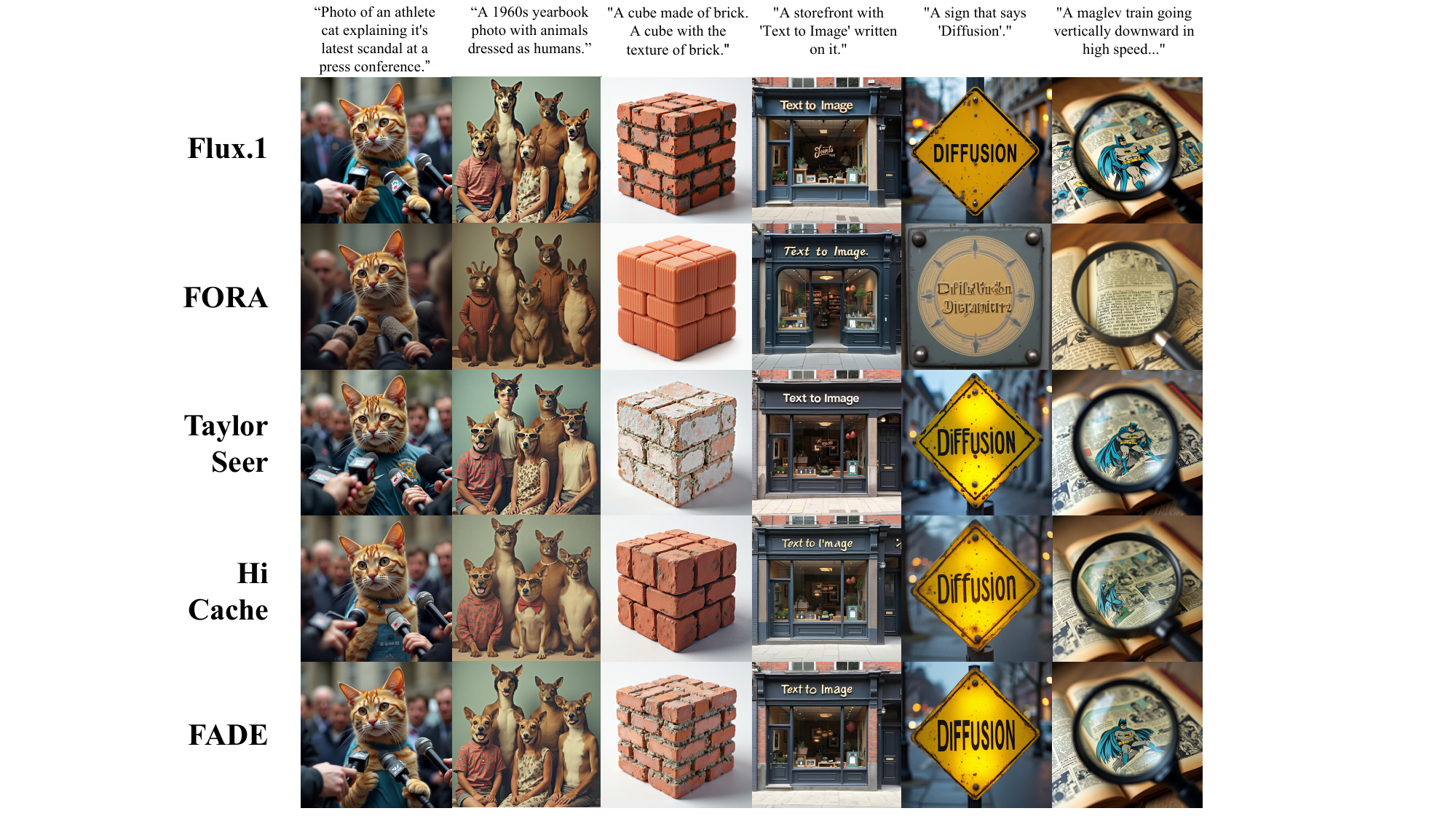}
    \caption{Visualization examples for different acceleration methods on Flux.}
    \label{fig:flux_viz_methods}
\end{figure}

\begin{table}[t]
  \centering
  \caption{Quantitative comparison of text-to-image generation on FLUX}
  \label{tab:main_flux_updated}
  \setlength{\tabcolsep}{1pt}
  \small    
  \resizebox{\textwidth}{!}{
  \begin{tabular}{lccccccccc}
    \toprule
    \textbf{Method}
    & \textbf{Latency(s)$\downarrow$}
    & \textbf{FLOPs(T)$\downarrow$}
    & \textbf{Speed$\uparrow$}
    & \textbf{Img. Reward$\uparrow$}
    & \textbf{CLIP Score$\uparrow$}
    & \textbf{Cycle Reward$\uparrow$}
    & \textbf{PSNR$\uparrow$}
    & \textbf{SSIM$\uparrow$}
    & \textbf{LPIPS$\downarrow$} \\
    \midrule
    FLUX.1 [dev] - 50 steps$^\dagger$
    & 17.18 & 3719.50 & 1.00$\times$ & 0.9898 & 27.4611 & 0.8800 & $\infty$ & 1.0000 & 0.0000 \\
    FLUX.1 [dev] - 25 steps
    & 8.80 & 1859.75 & 2.00$\times$ & 0.9375 & 27.3388 & 0.8466 & 19.2599 & 0.7749 & 0.3583 \\
    FLUX.1 [dev] - 20 steps
    & 7.12 & 1487.80 & 2.62$\times$ & 0.9387 & 27.2505 & 0.8466 & 18.1371 & 0.7484 & 0.4006 \\
    \midrule
    FORA ($\mathcal{N}$=5)~\cite{selvaraju2024FORAfastforwardcachingdiffusion}
    & 5.01 & 893.54 & 4.16$\times$ & 0.8253 & 27.2377 & 0.8353 & 15.9899 & 0.6677 & 0.5206 \\
    TaylorSeer ($\mathcal{N}$=5, m=1)~\cite{liu2025reusingforecastingacceleratingdiffusion}
    & 5.14 & 893.54 & 4.16$\times$ & 0.9919 & 27.5143 & 0.8540 & 18.4497 & 0.7424 & 0.4146 \\
    HiCache ($\mathcal{N}$=5, m=1)~\cite{feng2026hicachepluginscaledhermiteupgrade}
    & 5.12 & 893.54 & 4.16$\times$ & 0.9350 & 27.4034 & 0.8346 & 19.2595 & 0.7503 & 0.3911 \\
    \rowcolor{gray!10}
    \textbf{FADE$^\ddagger$ ($\mathcal{N}$=5, m=1)}
    & 5.04 & 893.54 & 4.16$\times$ & \textbf{1.0071} & \textbf{27.5724} & \textbf{0.8600} & \textbf{19.6683} & \textbf{0.7593} & \textbf{0.3815} \\
    \midrule
    FORA ($\mathcal{N}$=6)~\cite{selvaraju2024FORAfastforwardcachingdiffusion}
    & 4.40 & 744.81 & 4.99$\times$ & 0.7836 & 27.1541 & 0.8087 & 15.5290 & 0.6663 & 0.5253 \\
    TaylorSeer ($\mathcal{N}$=6, m=1)~\cite{liu2025reusingforecastingacceleratingdiffusion}
    & 4.57 & 744.81 & 4.99$\times$ & 0.9850 & 27.5199 & 0.8480 & 17.5396 & 0.7078 & 0.4657 \\
    HiCache ($\mathcal{N}$=6, m=1)~\cite{feng2026hicachepluginscaledhermiteupgrade}
    & 4.55 & 744.81 & 4.99$\times$ & 0.9427 & 27.6898 & 0.8539 & 18.5159 & 0.7155 & 0.4503 \\
    HiCache ($\mathcal{N}$=6, m=2)~\cite{feng2026hicachepluginscaledhermiteupgrade}
    & 5.03 & 744.81 & 4.99$\times$ & 0.9892 & 27.6586 & 0.8428 & 18.5044 & 0.7268 & 0.4316 \\
    \rowcolor{gray!10}
    \textbf{FADE$^\ddagger$ ($\mathcal{N}$=6, m=1)}
    & 4.43 & 744.81 & 4.99$\times$ & \textbf{1.0019}  & \textbf{27.7060} & \textbf{0.8706} & \textbf{18.7114} & \textbf{0.7320} & \textbf{0.4207} \\
    \midrule
    FORA ($\mathcal{N}$=7)~\cite{selvaraju2024FORAfastforwardcachingdiffusion}
    & 4.10 & 670.44 & 5.55$\times$ & 0.7289 & 27.0135 & 0.7962 & 15.5290 & 0.6548 & 0.5437 \\
    TaylorSeer ($\mathcal{N}$=7, m=1)~\cite{liu2025reusingforecastingacceleratingdiffusion}
    & 4.31 & 670.44 & 5.55$\times$ & 0.9420 & 27.4039 & 0.7970 & 16.8563 & 0.6844 & 0.5062 \\
    HiCache ($\mathcal{N}$=7, m=1)~\cite{feng2026hicachepluginscaledhermiteupgrade}
    & 4.20 & 670.44 & 5.55$\times$ & 0.9191 & 27.6949 & 0.8494 & \textbf{18.2778} & 0.7032 & 0.4708 \\
    HiCache ($\mathcal{N}$=7, m=2)~\cite{feng2026hicachepluginscaledhermiteupgrade}
    & 4.72 & 670.44 & 5.55$\times$ & 0.9881 & 27.5651 & 0.8384 & 17.9911 & 0.7098 & 0.4592 \\
    \rowcolor{gray!10}
    \textbf{FADE$^\ddagger$ ($\mathcal{N}$=7, m=1)}
    & 4.12 & 670.44 & 5.55$\times$ & 0.9861 & \textbf{27.8089} & \textbf{0.8621} & 18.1925 & \textbf{0.7137} & \textbf{0.4491} \\
    \textbf{FADE$^\ddagger$ ($\mathcal{N}$=7, m=2)}
    & 4.25 & 670.44 & 5.55$\times$ & \textbf{0.9914} & 27.7372 & 0.8592 & 17.7636 & 0.7106 & 0.4561 \\
    \bottomrule
  \end{tabular}
  }
\end{table}
\paragraph{Quantitative Analysis}
Table~\ref{tab:main_flux_updated} reports the quantitative results on FLUX. FADE achieves a strong quality--efficiency trade-off across acceleration settings, consistently outperforming existing baselines on semantic alignment and structural fidelity at standard configurations ($\mathcal{N}\in\{5,6\}$). Under a more aggressive skip interval, FADE still maintains competitive generation quality with negligible latency overhead, demonstrating the robustness of fractional forecasting.
\paragraph{Qualitative comparison.}
As illustrated in Figure~\ref{fig:flux_viz_methods}, FADE
maintains close visual consistency with the original FLUX.1
baseline in text rendering, texture, and scene composition.
It preserves the ``Text to Image'' storefront and the
``Diffusion'' sign, whereas
FORA~\cite{selvaraju2024FORAfastforwardcachingdiffusion}
alters the sign geometry and produces unclear lettering.
For the brick cube, FADE retains the reddish material and
mortar details, while
TaylorSeer~\cite{liu2025reusingforecastingacceleratingdiffusion}
produces a pale stone-like appearance, FORA generates
smoother surfaces, and
HiCache~\cite{feng2026hicachepluginscaledhermiteupgrade}
changes the brick arrangement.
FADE also preserves the cat press-conference composition
and animal yearbook scene, while TaylorSeer introduces
sunglasses and more human-like bodies in the latter.
In the magnifying-glass example, FADE retains the train
imagery within the lens, whereas FORA replaces it with
text-like patterns.

\subsection{Text-to-Video Generation}
\begin{table}[t]
  \centering
  \caption{Quantitative comparison of text-to-video generation on HunyuanVideo.}
  \label{tab:t2v_generation}
  \footnotesize
  \setlength{\tabcolsep}{3pt}
  \resizebox{\columnwidth}{!}{
  \begin{tabular}{lcccc}
    \toprule
    \textbf{Method} & \textbf{Lat. (s)$\downarrow$} & \textbf{FLOPs (T)$\downarrow$} & \textbf{Spd.$\uparrow$} & \textbf{VBench (\%)$\uparrow$} \\
    \midrule
    Original (50 steps) & 190.2 & 29773 & 1.00$\times$ & 80.68 \\
    DDIM (22\% steps)~\cite{song2022denoisingdiffusionimplicitmodels} & 42.7 & 6550 & 4.55$\times$ & 78.73 \\
    \midrule
    FORA (N=6)~\cite{selvaraju2024FORAfastforwardcachingdiffusion} & 37.1 & 5359 & 5.56$\times$ & 78.86 \\
    TaylorSeer (N=6, m=1)~\cite{liu2025reusingforecastingacceleratingdiffusion} & 37.8 & 5359 & 5.56$\times$ & 79.87 \\
    HiCache (N=6, m=1)~\cite{feng2026hicachepluginscaledhermiteupgrade} & 37.8 & 5359 & 5.56$\times$ & 79.93 \\
    \textbf{FADE (N=6, m=1)} & 37.9 & 5359 & 5.56$\times$ & \textbf{80.12} \\
    \bottomrule
  \end{tabular}
  }
  \vspace{-0.8em}
\end{table}
\begin{figure}[t]
    \centering
    \includegraphics[width=0.85\linewidth]{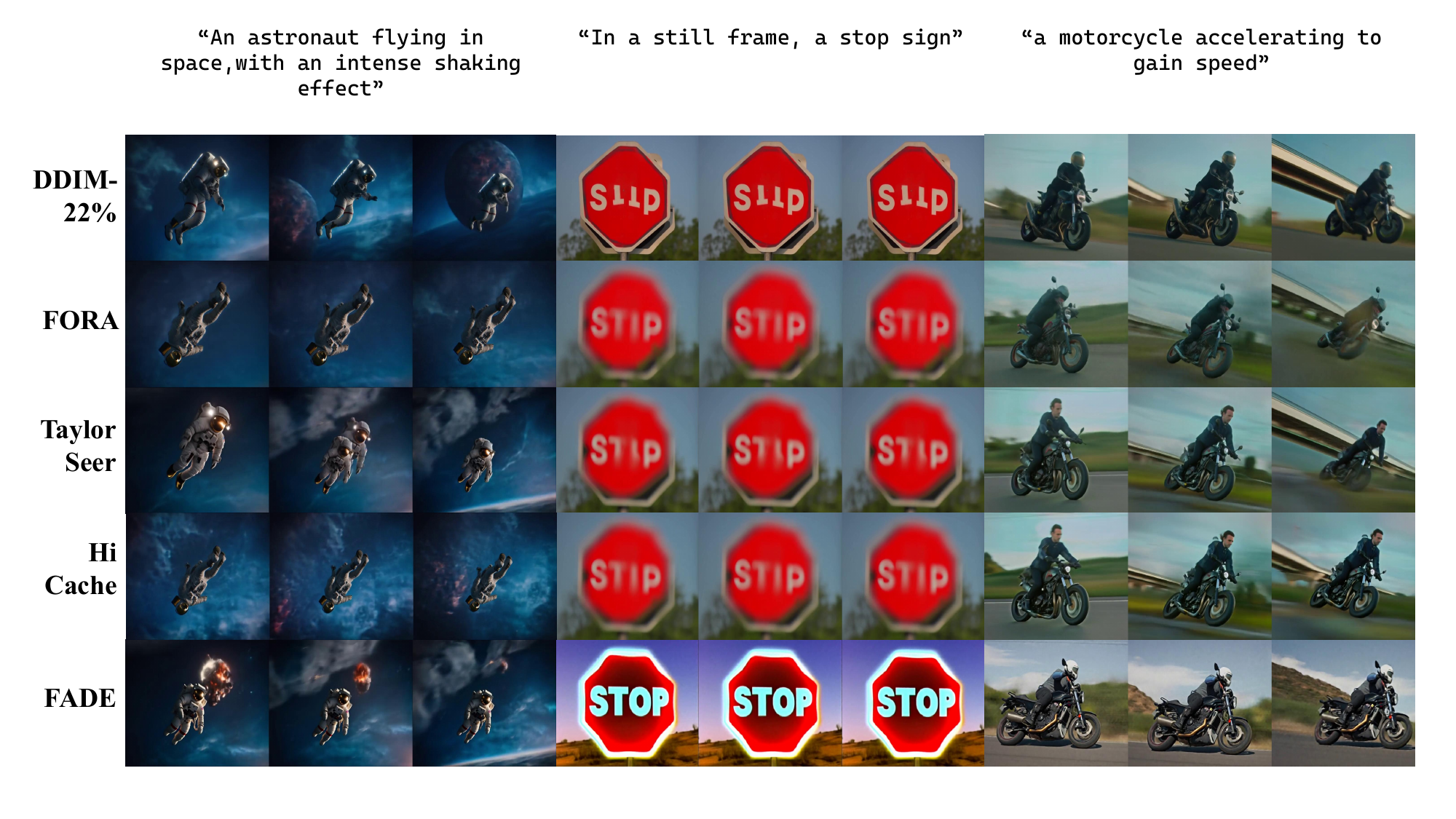}
    \caption{\textbf{Qualitative comparison on text-to-video generation.}}
    \label{fig:video_comparison}
\end{figure}
\paragraph{Quantitative Analysis.}
Table~\ref{tab:t2v_generation} reports the quantitative results
on HunyuanVideo. Under the same FLOPs budget and acceleration
ratio, FADE achieves competitive latency and the best VBench
score, indicating that fractional forecasting remains effective
for text-to-video generation.

\paragraph{Qualitative Comparison.}
As shown in Fig.~\ref{fig:video_comparison}, FADE preserves
clearer text and structure in the static case, correctly
rendering the word ``STOP'', while better maintaining motion
semantics and scene context in dynamic prompts. In particular,
FADE better captures the intended temporal behavior and
preserves prompt-specific motion cues, rather than merely
maintaining frame-level appearance. In contrast, the baselines
either miss the intended intense shaking effect in the
astronaut case or introduce incorrect motion directions and
off-road backgrounds in the motorcycle case.

\subsection{Ablation Study}

\begin{table*}[t]
  \centering
  \begin{minipage}{0.85\textwidth}
  \centering
  \begin{minipage}[t]{0.485\linewidth}
    \centering
    \captionof{table}{Ablation on DiT-XL/2.}
    \label{tab:ablation_dit_compact}
    \setlength{\tabcolsep}{2.6pt}
    \scriptsize
    \resizebox{\linewidth}{!}{
    \begin{tabular}{lccc}
      \toprule
      \textbf{Method} & \textbf{Speed$\uparrow$} & \textbf{FID$\downarrow$} & \textbf{sFID$\downarrow$} \\
      \midrule
      DDIM-50 & 1.00$\times$ & 2.29 & 4.29 \\
      DDIM-10 & 5.00$\times$ & 12.17 & 11.24 \\ 
      HiCache & 4.98$\times$ & 3.09 & 6.30 \\
      \midrule
      FADE ($\gamma=1$) & 4.98$\times$ & 3.17 & 6.36 \\
      FADE (fixed $\gamma=0.85$) & 4.98$\times$ & 3.06 & 6.05 \\
      FADE (poly-fit $\gamma(t)$) & 4.98$\times$ & 3.00 & 6.19 \\
      FADE (LUT + MSE) & 4.98$\times$ & 2.97 & 5.91 \\
      \rowcolor{gray!10}
      \textbf{FADE (LUT + Dir.)} & \textbf{4.98$\times$} & \textbf{2.87} & \textbf{5.73} \\
      \bottomrule
    \end{tabular}
    }
  \end{minipage}%
  \hfill
  \begin{minipage}[t]{0.485\linewidth}
    \centering
    \captionof{table}{Ablation on FLUX.1.}
    \label{tab:ablation_flux_compact}
    \setlength{\tabcolsep}{2.6pt}
    \scriptsize
    \resizebox{\linewidth}{!}{
    \begin{tabular}{lccc}
      \toprule
      \textbf{Method} & \textbf{Speed$\uparrow$} & \textbf{IR$\uparrow$} & \textbf{CLIP$\uparrow$} \\
      \midrule
      FADE ($|\mathcal{D}_{\text{calib}}|=1$) & 4.99$\times$ & 0.9963 & 27.6781 \\
      FADE ($|\mathcal{D}_{\text{calib}}|=5$)  & 4.99$\times$ & 1.0015 & 27.7454 \\
      \rowcolor{gray!10}
      FADE ($|\mathcal{D}_{\text{calib}}|=10$) & 4.99$\times$ & \textbf{1.0019} & 27.7060 \\
      FADE ($|\mathcal{D}_{\text{calib}}|=20$) & 4.99$\times$ & 1.0008 & 27.7256 \\
      \midrule
      FADE ($\lambda=1.0$) & 4.16$\times$ & 0.9974 & 27.4618 \\
      FADE ($\lambda=1.2$) & 4.16$\times$ & 0.9994 & \textbf{27.6198} \\
      \rowcolor{gray!10}
      \textbf{FADE ($\lambda=1.3$)} & \textbf{4.16$\times$} & \textbf{1.0071} & 27.5724 \\
      FADE ($\lambda=1.4$) & 4.16$\times$ & 0.9858 & 27.5050 \\
      \bottomrule
    \end{tabular}
    }
  \end{minipage}
  \end{minipage}
\end{table*}

On DiT-XL/2, we compare several strategies for determining the fractional order, including fixed global tuning, polynomial fitting over denoising progress, and the proposed stage-aware LUT retrieval. We also compare MSE-based and directional calibration objectives. As shown in Table~\ref{tab:ablation_dit_compact}, the full FADE design achieves the best performance, suggesting that the optimal fractional dynamics cannot be well captured by either a static setting or a simple smooth fit, and that directional calibration provides a more effective signal for stable offline calibration.

On FLUX.1, we study two practical aspects of offline calibration: the calibration set size and the global refinement factor $\lambda$. Table~\ref{tab:ablation_flux_compact} shows that FADE remains effective with a small calibration set, saturating around $|\mathcal{D}_{\text{calib}}|=10$. A moderate refinement factor gives the best overall performance, with $\lambda=1.3$ achieving the best ImageReward and $\lambda=1.2$ the best CLIPScore. This indicates that $\lambda$ is a useful lightweight correction, though its optimal value depends on the prediction interval and sampling setting. Additional ablations are provided in Appendix~\ref{sec:addition_ablation}.

We further report the practical cost of offline calibration
and LUT retrieval. Using 10 calibration samples, end-to-end
offline calibration takes approximately 50\,s, 550\,s, and
2{,}000\,s for DiT-XL/2, FLUX.1, and HunyuanVideo,
respectively, on their respective evaluation hardware.
The LUT has a small storage footprint, occupying only
44.8\,kB for DiT-XL/2. The accumulated LUT-retrieval time
per generation is 0.14\,ms, 0.57\,ms, and 0.61\,ms,
respectively.

\section{Conclusion and Future Work}

This work revisits training-free DiT acceleration from the
perspective of fractional dynamics. Motivated by the limitations
of cache-based methods under complex stage-dependent latent
dynamics, we view DiT feature evolution through the lens of
anomalous diffusion and formulate FADE using a
Mittag--Leffler-based fractional operator. With an
offline-profiled LUT, FADE enables adaptive forecasting at
inference time with $\mathcal{O}(1)$ retrieval and no extra
runtime overhead. Experiments across diverse architectures
and modalities show that FADE achieves a stronger
quality--efficiency trade-off than existing baselines while
remaining robust under large prediction distances.

FADE relies on manually designed calibration, and its
Mittag--Leffler operator is chosen primarily for simplicity.
Future work will explore alternative fractional operators
and order parameterizations to improve forecasting accuracy,
stability, and the quality--efficiency trade-off across
denoising stages, network locations, and skip intervals.

\begin{ack}
This work was supported by National Natural Science Foundation of China under Grant No.62376172.
\end{ack}

\bibliographystyle{plainnat}
\bibliography{main}

@inproceedings{ho2020denoisingdiffusionprobabilisticmodels,
 author = {Ho, Jonathan and Jain, Ajay and Abbeel, Pieter},
 booktitle = {Advances in Neural Information Processing Systems},
 editor = {H. Larochelle and M. Ranzato and R. Hadsell and M.F. Balcan and H. Lin},
 pages = {6840--6851},
 publisher = {Curran Associates, Inc.},
 title = {Denoising Diffusion Probabilistic Models},
 url = {https://proceedings.neurips.cc/paper_files/paper/2020/file/4c5bcfec8584af0d967f1ab10179ca4b-Paper.pdf},
 volume = {33},
 year = {2020}
}

@inproceedings{peebles2023scalablediffusionmodelstransformers,
  title={Scalable diffusion models with transformers},
  author={Peebles, William and Xie, Saining},
  booktitle={Proceedings of the IEEE/CVF international conference on computer vision},
  pages={4195--4205},
  year={2023}
}

@inproceedings{ma2023deepcacheacceleratingdiffusionmodels,
  title={Deepcache: Accelerating diffusion models for free},
  author={Ma, Xinyin and Fang, Gongfan and Wang, Xinchao},
  booktitle={Proceedings of the IEEE/CVF conference on computer vision and pattern recognition},
  pages={15762--15772},
  year={2024}
}

@misc{selvaraju2024FORAfastforwardcachingdiffusion,
      title={FORA: Fast-Forward Caching in Diffusion Transformer Acceleration}, 
      author={Pratheba Selvaraju and Tianyu Ding and Tianyi Chen and Ilya Zharkov and Luming Liang},
      year={2024},
      eprint={2407.01425},
      archivePrefix={arXiv},
      primaryClass={cs.CV},
      url={https://arxiv.org/abs/2407.01425}, 
}

@inproceedings{liu2025reusingforecastingacceleratingdiffusion,
  title={From reusing to forecasting: Accelerating diffusion models with taylorseers},
  author={Liu, Jiacheng and Zou, Chang and Lyu, Yuanhuiyi and Chen, Junjie and Zhang, Linfeng},
  booktitle={Proceedings of the IEEE/CVF International Conference on Computer Vision},
  pages={15853--15863},
  year={2025}
}

@inproceedings{
shang2023ptqdm,
title={Post-training Quantization on Diffusion Models},
author={Yuzhang Shang and Zhihang Yuan and Bin Xie and Bingzhe Wu and Yan Yan},
booktitle={CVPR},
year={2023}
}

@article{liu2023audioldm,
  title={Audioldm: Text-to-audio generation with latent diffusion models},
  author={Liu, Haohe and Chen, Zehua and Yuan, Yi and Mei, Xinhao and Liu, Xubo and Mandic, Danilo and Wang, Wenwu and Plumbley, Mark D},
  journal={arXiv preprint arXiv:2301.12503},
  year={2023}
}

@article{lu2022dpmsolverfastodesolver,
  title={Dpm-solver: A fast ode solver for diffusion probabilistic model sampling in around 10 steps},
  author={Lu, Cheng and Zhou, Yuhao and Bao, Fan and Chen, Jianfei and Li, Chongxuan and Zhu, Jun},
  journal={Advances in neural information processing systems},
  volume={35},
  pages={5775--5787},
  year={2022}
}

@inproceedings{huang2023vbenchcomprehensivebenchmarksuite,
  title={Vbench: Comprehensive benchmark suite for video generative models},
  author={Huang, Ziqi and He, Yinan and Yu, Jiashuo and Zhang, Fan and Si, Chenyang and Jiang, Yuming and Zhang, Yuanhan and Wu, Tianxing and Jin, Qingyang and Chanpaisit, Nattapol and others},
  booktitle={Proceedings of the IEEE/CVF Conference on Computer Vision and Pattern Recognition},
  pages={21807--21818},
  year={2024}
}

@article{kong2024hunyuanvideo,
  title={Hunyuanvideo: A systematic framework for large video generative models},
  author={Kong, Weijie and Tian, Qi and Zhang, Zijian and Min, Rox and Dai, Zuozhuo and Zhou, Jin and Xiong, Jiangfeng and Li, Xin and Wu, Bo and Zhang, Jianwei and others},
  journal={arXiv preprint arXiv:2412.03603},
  year={2024}
}

@article{oliveira2019anomalous,
  title={Anomalous diffusion: A basic mechanism for the evolution of inhomogeneous systems},
  author={Oliveira, Fernando A and Ferreira, Rogelma MS and Lapas, Luciano C and Vainstein, Mendeli H},
  journal={Frontiers in Physics},
  volume={7},
  pages={18},
  year={2019},
  publisher={Frontiers Media SA}
}

@misc{chen2023pixartalphafasttrainingdiffusion,
      title={PixArt-$\alpha$: Fast Training of Diffusion Transformer for Photorealistic Text-to-Image Synthesis}, 
      author={Junsong Chen and Jincheng Yu and Chongjian Ge and Lewei Yao and Enze Xie and Yue Wu and Zhongdao Wang and James Kwok and Ping Luo and Huchuan Lu and Zhenguo Li},
      year={2023},
      eprint={2310.00426},
      archivePrefix={arXiv},
      primaryClass={cs.CV},
      url={https://arxiv.org/abs/2310.00426}, 
}

@misc{song2022denoisingdiffusionimplicitmodels,
      title={Denoising Diffusion Implicit Models}, 
      author={Jiaming Song and Chenlin Meng and Stefano Ermon},
      year={2022},
      eprint={2010.02502},
      archivePrefix={arXiv},
      primaryClass={cs.LG},
      url={https://arxiv.org/abs/2010.02502}, 
}

@article{Lu_2025,
   title={DPM-Solver++: Fast Solver for Guided Sampling of Diffusion Probabilistic Models},
   volume={22},
   ISSN={2731-5398},
   url={http://dx.doi.org/10.1007/s11633-025-1562-4},
   DOI={10.1007/s11633-025-1562-4},
   number={4},
   journal={Machine Intelligence Research},
   publisher={Springer Science and Business Media LLC},
   author={Lu, Cheng and Zhou, Yuhao and Bao, Fan and Chen, Jianfei and Li, Chongxuan and Zhu, Jun},
   year={2025},
   month=jun, pages={730–751} }

@misc{liu2022flowstraightfastlearning,
      title={Flow Straight and Fast: Learning to Generate and Transfer Data with Rectified Flow}, 
      author={Xingchao Liu and Chengyue Gong and Qiang Liu},
      year={2022},
      eprint={2209.03003},
      archivePrefix={arXiv},
      primaryClass={cs.LG},
      url={https://arxiv.org/abs/2209.03003}, 
}

@misc{song2023consistencymodels,
      title={Consistency Models}, 
      author={Yang Song and Prafulla Dhariwal and Mark Chen and Ilya Sutskever},
      year={2023},
      eprint={2303.01469},
      archivePrefix={arXiv},
      primaryClass={cs.LG},
      url={https://arxiv.org/abs/2303.01469}, 
}

@misc{fang2023structuralpruningdiffusionmodels,
      title={Structural Pruning for Diffusion Models}, 
      author={Gongfan Fang and Xinyin Ma and Xinchao Wang},
      year={2023},
      eprint={2305.10924},
      archivePrefix={arXiv},
      primaryClass={cs.LG},
      url={https://arxiv.org/abs/2305.10924}, 
}

@inproceedings{li2023qdiffusionquantizingdiffusionmodels,
  title={Q-diffusion: Quantizing diffusion models},
  author={Li, Xiuyu and Liu, Yijiang and Lian, Long and Yang, Huanrui and Dong, Zhen and Kang, Daniel and Zhang, Shanghang and Keutzer, Kurt},
  booktitle={Proceedings of the IEEE/CVF International Conference on Computer Vision},
  pages={17535--17545},
  year={2023}
}

@misc{zhang2024tokenpruningcachingbetter,
      title={Token Pruning for Caching Better: 9 Times Acceleration on Stable Diffusion for Free}, 
      author={Evelyn Zhang and Bang Xiao and Jiayi Tang and Qianli Ma and Chang Zou and Xuefei Ning and Xuming Hu and Linfeng Zhang},
      year={2024},
      eprint={2501.00375},
      archivePrefix={arXiv},
      primaryClass={cs.CV},
      url={https://arxiv.org/abs/2501.00375}, 
}

@inproceedings{bolya2023tokenmergingfaststable,
  title={Token merging for fast stable diffusion},
  author={Bolya, Daniel and Hoffman, Judy},
  booktitle={Proceedings of the IEEE/CVF conference on computer vision and pattern recognition},
  pages={4599--4603},
  year={2023}
}

@misc{salimans2022progressivedistillationfastsampling,
      title={Progressive Distillation for Fast Sampling of Diffusion Models}, 
      author={Tim Salimans and Jonathan Ho},
      year={2022},
      eprint={2202.00512},
      archivePrefix={arXiv},
      primaryClass={cs.LG},
      url={https://arxiv.org/abs/2202.00512}, 
}

@inproceedings{ma2024sitexploringflowdiffusionbased,
  title={Sit: Exploring flow and diffusion-based generative models with scalable interpolant transformers},
  author={Ma, Nanye and Goldstein, Mark and Albergo, Michael S and Boffi, Nicholas M and Vanden-Eijnden, Eric and Xie, Saining},
  booktitle={European Conference on Computer Vision},
  pages={23--40},
  year={2024},
  organization={Springer}
}

@INPROCEEDINGS{deng2009imagenet,
  author={Deng, Jia and Dong, Wei and Socher, Richard and Li, Li-Jia and Kai Li and Li Fei-Fei},
  booktitle={2009 IEEE Conference on Computer Vision and Pattern Recognition},
  title={ImageNet: A large-scale hierarchical image database},
  year={2009},
  pages={248-255},
  doi={10.1109/CVPR.2009.5206848}
}

@inproceedings{zheng2025forecastcalibratefeaturecaching,
  author       = {Shikang Zheng and Liang Feng and Xinyu Wang and Qinming Zhou and Peiliang Cai and Chang Zou and Jiacheng Liu and Yuqi Lin and Junjie Chen and Yue Ma and Linfeng Zhang},
  title        = {Forecast then Calibrate: Feature Caching as {ODE} for Efficient Diffusion Transformers},
  booktitle    = {Proceedings of the AAAI Conference on Artificial Intelligence},
  year         = {2026},

  url          = {https://arxiv.org/abs/2508.16211},
  eprint       = {2508.16211},
    
}

@inproceedings{feng2026hicachepluginscaledhermiteupgrade,
  title={HiCache: A Plug-in Scaled-Hermite Upgrade for Taylor-Style Cache-then-Forecast Diffusion Acceleration},
  author={Feng, Liang and Zheng, Shikang and Liu, Jiacheng and Lin, Yuqi and Zhou, Qinming and Cai, Peiliang and Wang, Xinyu and Chen, Junjie and Zou, Chang and Ma, Yue and Zhang, Linfeng},
  booktitle={International Conference on Learning Representations (ICLR)},
  year={2026}
}

@misc{zou2025acceleratingdiffusiontransformerstokenwise,
      title={Accelerating Diffusion Transformers with Token-wise Feature Caching}, 
      author={Chang Zou and Xuyang Liu and Ting Liu and Siteng Huang and Linfeng Zhang},
      year={2025},
      eprint={2410.05317},
      archivePrefix={arXiv},
      primaryClass={cs.LG},
      url={https://arxiv.org/abs/2410.05317}, 
}

@article{zhao2023unipcunifiedpredictorcorrectorframework,
  title={Unipc: A unified predictor-corrector framework for fast sampling of diffusion models},
  author={Zhao, Wenliang and Bai, Lujia and Rao, Yongming and Zhou, Jie and Lu, Jiwen},
  journal={Advances in Neural Information Processing Systems},
  volume={36},
  pages={49842--49869},
  year={2023}
}

@misc{luo2023latentconsistencymodelssynthesizing,
      title={Latent Consistency Models: Synthesizing High-Resolution Images with Few-Step Inference}, 
      author={Simian Luo and Yiqin Tan and Longbo Huang and Jian Li and Hang Zhao},
      year={2023},
      eprint={2310.04378},
      archivePrefix={arXiv},
      primaryClass={cs.CV},
      url={https://arxiv.org/abs/2310.04378}, 
}

@inproceedings{
      yuan2024ditfastattn,
      title={Di{TF}astAttn: Attention Compression for Diffusion Transformer Models},
      author={Zhihang Yuan and Hanling Zhang and Lu Pu and Xuefei Ning and Linfeng Zhang and Tianchen Zhao and Shengen Yan and Guohao Dai and Yu Wang},
      booktitle={The Thirty-eighth Annual Conference on Neural Information Processing Systems},
      year={2024},
      url={https://openreview.net/forum?id=51HQpkQy3t}
}

@misc{cai2025hidreami1highefficientimagegenerative,
      title={HiDream-I1: A High-Efficient Image Generative Foundation Model with Sparse Diffusion Transformer}, 
      author={Qi Cai and Jingwen Chen and Yang Chen and Yehao Li and Fuchen Long and Yingwei Pan and Zhaofan Qiu and Yiheng Zhang and Fengbin Gao and Peihan Xu and Yimeng Wang and Kai Yu and Wenxuan Chen and Ziwei Feng and Zijian Gong and Jianzhuang Pan and Yi Peng and Rui Tian and Siyu Wang and Bo Zhao and Ting Yao and Tao Mei},
      year={2025},
      eprint={2505.22705},
      archivePrefix={arXiv},
      primaryClass={cs.CV},
      url={https://arxiv.org/abs/2505.22705}, 
}

@article{wan2025,
      title={Wan: Open and Advanced Large-Scale Video Generative Models}, 
      author={Team Wan and Ang Wang and Baole Ai and Bin Wen and Chaojie Mao and Chen-Wei Xie and Di Chen and Feiwu Yu and Haiming Zhao and Jianxiao Yang and Jianyuan Zeng and Jiayu Wang and Jingfeng Zhang and Jingren Zhou and Jinkai Wang and Jixuan Chen and Kai Zhu and Kang Zhao and Keyu Yan and Lianghua Huang and Mengyang Feng and Ningyi Zhang and Pandeng Li and Pingyu Wu and Ruihang Chu and Ruili Feng and Shiwei Zhang and Siyang Sun and Tao Fang and Tianxing Wang and Tianyi Gui and Tingyu Weng and Tong Shen and Wei Lin and Wei Wang and Wei Wang and Wenmeng Zhou and Wente Wang and Wenting Shen and Wenyuan Yu and Xianzhong Shi and Xiaoming Huang and Xin Xu and Yan Kou and Yangyu Lv and Yifei Li and Yijing Liu and Yiming Wang and Yingya Zhang and Yitong Huang and Yong Li and You Wu and Yu Liu and Yulin Pan and Yun Zheng and Yuntao Hong and Yupeng Shi and Yutong Feng and Zeyinzi Jiang and Zhen Han and Zhi-Fan Wu and Ziyu Liu},
      journal = {arXiv preprint arXiv:2503.20314},
      year={2025}
}

@misc{kong2021diffwaveversatilediffusionmodel,
      title={DiffWave: A Versatile Diffusion Model for Audio Synthesis}, 
      author={Zhifeng Kong and Wei Ping and Jiaji Huang and Kexin Zhao and Bryan Catanzaro},
      year={2021},
      eprint={2009.09761},
      archivePrefix={arXiv},
      primaryClass={eess.AS},
      url={https://arxiv.org/abs/2009.09761}, 
}

@misc{saharia2022photorealistictexttoimagediffusionmodels,
      title={Photorealistic Text-to-Image Diffusion Models with Deep Language Understanding}, 
      author={Chitwan Saharia and William Chan and Saurabh Saxena and Lala Li and Jay Whang and Emily Denton and Seyed Kamyar Seyed Ghasemipour and Burcu Karagol Ayan and S. Sara Mahdavi and Rapha Gontijo Lopes and Tim Salimans and Jonathan Ho and David J Fleet and Mohammad Norouzi},
      year={2022},
      eprint={2205.11487},
      archivePrefix={arXiv},
      primaryClass={cs.CV},
      url={https://arxiv.org/abs/2205.11487}, 
}

@misc{labs2025flux1kontextflowmatching,
      title={FLUX.1 Kontext: Flow Matching for In-Context Image Generation and Editing in Latent Space},
      author={Black Forest Labs and Stephen Batifol and Andreas Blattmann and Frederic Boesel and Saksham Consul and Cyril Diagne and Tim Dockhorn and Jack English and Zion English and Patrick Esser and Sumith Kulal and Kyle Lacey and Yam Levi and Cheng Li and Dominik Lorenz and Jonas Müller and Dustin Podell and Robin Rombach and Harry Saini and Axel Sauer and Luke Smith},
      year={2025},
      eprint={2506.15742},
      archivePrefix={arXiv},
      primaryClass={cs.GR},
      url={https://arxiv.org/abs/2506.15742},
}

@misc{flux2024,
    author={Black Forest Labs},
    title={FLUX},
    year={2024},
    howpublished={\url{https://github.com/black-forest-labs/flux}},
}

@inproceedings{xu2023imagereward,
  title={ImageReward: learning and evaluating human preferences for text-to-image generation},
  author={Xu, Jiazheng and Liu, Xiao and Wu, Yuchen and Tong, Yuxuan and Li, Qinkai and Ding, Ming and Tang, Jie and Dong, Yuxiao},
  booktitle={Proceedings of the 37th International Conference on Neural Information Processing Systems},
  pages={15903--15935},
  year={2023}
}

@misc{yang2026bracetamingsharpirregularities,
      title={BRACE: Taming Sharp Irregularities via Barycentric Rational Forecasting for Fast Diffusion Transformers Inference}, 
      author={Jinlong Yang and Jinke Wu and Lizilin and Yao Zhou},
      year={2026},
      eprint={2608.07572},
      archivePrefix={arXiv},
      primaryClass={cs.CV},
      url={https://arxiv.org/abs/2608.07572}, 
}

@misc{hessel2022clipscorereferencefreeevaluationmetric,
      title={CLIPScore: A Reference-free Evaluation Metric for Image Captioning}, 
      author={Jack Hessel and Ari Holtzman and Maxwell Forbes and Ronan Le Bras and Yejin Choi},
      year={2022},
      eprint={2104.08718},
      archivePrefix={arXiv},
      primaryClass={cs.CV},
      url={https://arxiv.org/abs/2104.08718}, 
}

@InProceedings{Bahng_2025_ICCV,
    author    = {Bahng, Hyojin and Chan, Caroline and Durand, Fredo and Isola, Phillip},
    title     = {Cycle Consistency as Reward: Learning Image-Text Alignment without Human Preferences},
    booktitle = {Proceedings of the IEEE/CVF International Conference on Computer Vision (ICCV)},
    month     = {October},
    year      = {2025},
    pages     = {22934-22946}
}

@INPROCEEDINGS{PSNRSSIM,
  author={Horé, Alain and Ziou, Djemel},
  booktitle={2010 20th International Conference on Pattern Recognition}, 
  title={Image Quality Metrics: PSNR vs. SSIM}, 
  year={2010},
  volume={},
  number={},
  pages={2366-2369},
  doi={10.1109/ICPR.2010.579}}

@InProceedings{LPIPsZhang_2018_CVPR,
author = {Zhang, Richard and Isola, Phillip and Efros, Alexei A. and Shechtman, Eli and Wang, Oliver},
title = {The Unreasonable Effectiveness of Deep Features as a Perceptual Metric},
booktitle = {Proceedings of the IEEE Conference on Computer Vision and Pattern Recognition (CVPR)},
month = {June},
year = {2018}
}

@misc{fid,
      title={GANs Trained by a Two Time-Scale Update Rule Converge to a Local Nash Equilibrium}, 
      author={Martin Heusel and Hubert Ramsauer and Thomas Unterthiner and Bernhard Nessler and Sepp Hochreiter},
      year={2018},
      eprint={1706.08500},
      archivePrefix={arXiv},
      primaryClass={cs.LG},
      url={https://arxiv.org/abs/1706.08500}, 
}

@ARTICLE{11711089,
  author={Zhou, Yao and Yi, Zhang and Yen, Gary G.},
  journal={IEEE Transactions on Cybernetics}, 
  title={Multiobjective Evolutionary Diffusion Transformer Acceleration}, 
  year={2026},
  volume={56},
  number={11},
  pages={6085-6096},
  doi={10.1109/TCYB.2026.3731615}}

@article{METZLER20001,
title = {The random walk's guide to anomalous diffusion: a fractional dynamics approach},
journal = {Physics Reports},
volume = {339},
number = {1},
pages = {1-77},
year = {2000},
issn = {0370-1573},
doi = {https://doi.org/10.1016/S0370-1573(00)00070-3},
url = {https://www.sciencedirect.com/science/article/pii/S0370157300000703},
author = {Ralf Metzler and Joseph Klafter}
}

@article{kalatzis2020variational,
  title={Variational autoencoders with riemannian brownian motion priors},
  author={Kalatzis, Dimitris and Eklund, David and Arvanitidis, Georgios and Hauberg, S{\o}ren},
  journal={arXiv preprint arXiv:2002.05227},
  year={2020}
}

@article{lobashev2025hessian,
  title={Hessian geometry of latent space in generative models},
  author={Lobashev, Alexander and Guskov, Dmitry and Larchenko, Maria and Tamm, Mikhail},
  journal={arXiv preprint arXiv:2506.10632},
  year={2025}
}

@article{zhao2024real,
  title={Real-time video generation with pyramid attention broadcast},
  author={Zhao, Xuanlei and Jin, Xiaolong and Wang, Kai and You, Yang},
  journal={arXiv preprint arXiv:2408.12588},
  year={2024}
}

@article{zhang2024token,
  title={Token pruning for caching better: 9 times acceleration on stable diffusion for free},
  author={Zhang, Evelyn and Xiao, Bang and Tang, Jiayi and Ma, Qianli and Zou, Chang and Ning, Xuefei and Hu, Xuming and Zhang, Linfeng},
  journal={arXiv preprint arXiv:2501.00375},
  year={2024}
}

@article{lou2024token,
  title={Token caching for diffusion transformer acceleration},
  author={Lou, Jinming and Luo, Wenyang and Liu, Yufan and Li, Bing and Ding, Xinmiao and Hu, Weiming and Li, Yuming and Ma, Chenguang},
  journal={arXiv preprint arXiv:2409.18523},
  year={2024}
}

@article{yang2025lrq,
  title={LRQ-DiT: Log-Rotation Post-Training Quantization of Diffusion Transformers for Image and Video Generation},
  author={Yang, Lianwei and Lin, Haokun and Zhao, Tianchen and Wu, Yichen and Zhu, Hongyu and Xie, Ruiqi and Sun, Zhenan and Wang, Yu and Gu, Qingyi},
  journal={arXiv preprint arXiv:2508.03485},
  year={2025}
}

@inproceedings{liu2025timestep,
  title={Timestep Embedding Tells: It's Time to Cache for Video Diffusion Model},
  author={Liu, Feng and Zhang, Shiwei and Wang, Xiaofeng and Wei, Yujie and Qiu, Haonan and Zhao, Yuzhong and Zhang, Yingya and Ye, Qixiang and Wan, Fang},
  booktitle={Proceedings of the Computer Vision and Pattern Recognition Conference},
  pages={7353--7363},
  year={2025}
}

@article{peng2025ertacache,
  title={Ertacache: Error rectification and timesteps adjustment for efficient diffusion},
  author={Peng, Xurui and Yan, Chenqian and Liu, Hong and Ma, Rui and Chen, Fangmin and Wang, Xing and Wu, Zhihua and Liu, Songwei and Lin, Mingbao},
  journal={arXiv preprint arXiv:2508.21091},
  year={2025}
}

@inproceedings{kahatapitiya2025adaptive,
  title={Adaptive caching for faster video generation with diffusion transformers},
  author={Kahatapitiya, Kumara and Liu, Haozhe and He, Sen and Liu, Ding and Jia, Menglin and Zhang, Chenyang and Ryoo, Michael S and Xie, Tian},
  booktitle={Proceedings of the IEEE/CVF International Conference on Computer Vision},
  pages={15240--15252},
  year={2025}
}

\appendix
\section*{Technical appendices}

\section{Detailed Derivation of the FADE Predictive Formulation}
\label{sec:appendix_derivation}

In this section, we provide the rigorous step-by-step mathematical derivation detailing the transition from the classical integer-order extrapolation to our generalized Fractional Anomalous Dynamics Extrapolation (FADE) formulation.

\subsection{From Polynomial Extrapolation to the Exponential Shift Operator}

Based on operator calculus, the polynomial extrapolation in Eq.~\eqref{eq:taylor} can be interpreted as applying an exponential shift operator to the current feature representation. 
Let $\mathbf{F}_t = \mathbf{F}(\mathbf{x}^l_t)$ denote the feature at timestep $t$, and let $\delta$ denote the prediction distance from $t$ to $t-\delta$. 
We define the temporal derivative operator as
\begin{equation}
    \nabla_t = \frac{\partial}{\partial t}.
\end{equation}
For a sufficiently smooth continuous trajectory, the Taylor expansion of $\mathbf{F}_{t-\delta}$ around $t$ can be written as
\begin{equation}
    \mathbf{F}_{t-\delta}
    =
    \mathbf{F}_t
    +
    \sum_{i=1}^{\infty}
    \frac{\nabla_t^i \mathbf{F}_t}{i!}(-\delta)^i .
\end{equation}
Equivalently, by collecting the derivative operator terms, we obtain
\begin{equation}
    \mathbf{F}_{t-\delta}
    =
    \left(
    \sum_{i=0}^{\infty}
    \frac{(-\delta \nabla_t)^i}{i!}
    \right)
    \mathbf{F}_t .
\end{equation}
Since the infinite series corresponds to the Maclaurin expansion of the exponential function, this transformation can be compactly expressed as
\begin{equation}
    \mathbf{F}_{t-\delta}
    =
    e^{-\delta \nabla_t}\mathbf{F}_t .
\end{equation}
By defining
\begin{equation}
    z = -\delta \nabla_t ,
\end{equation}
the above relation becomes
\begin{equation}
    \mathbf{F}_{t-\delta} \approx e^z \mathbf{F}_t,
    \label{eq:appendix_operator_taylor}
\end{equation}
where the approximation emphasizes the practical setting in which the continuous operator is implemented through finite-order truncation and discrete feature differences.

In practical predictive caching, the infinite-dimensional operator is truncated to the $m$-th order:
\begin{equation}
    e^{-\delta \nabla_t}\mathbf{F}_t
    \approx
    \left(
    \sum_{i=0}^{m}
    \frac{(-\delta \nabla_t)^i}{i!}
    \right)
    \mathbf{F}_t
    =
    \mathbf{F}_t
    +
    \sum_{i=1}^{m}
    \frac{\nabla_t^i \mathbf{F}_t}{i!}(-\delta)^i .
\end{equation}
By approximating the continuous derivatives with discrete feature differences, i.e.,
\begin{equation}
    \Delta^i \mathbf{F}(\mathbf{x}^l_t)
    \approx
    \nabla_t^i \mathbf{F}_t ,
\end{equation}
we recover the polynomial forecasting form:
\begin{equation}
    \mathbf{F}_{\mathrm{pred}}(\mathbf{x}^l_{t-\delta})
    =
    \mathbf{F}(\mathbf{x}^l_t)
    +
    \sum_{i=1}^{m}
    \frac{\Delta^i \mathbf{F}(\mathbf{x}^l_t)}{i!}
    (-\delta)^i .
\end{equation}
Therefore, the commonly used $m$-th order polynomial extrapolation can be viewed as a truncated and discretized approximation of the exponential shift operator $e^{-\delta \nabla_t}$.

\subsection{Detailed Derivation of the FADE Predictive Formulation}
\label{subsec:appendix_fractional_derivation}

In this section, we provide a step-by-step derivation of the FADE predictive formulation, showing how the classical integer-order extrapolation can be generalized through a fractional Mittag--Leffler operator.

\paragraph{Operator Equivalence.}
We first revisit the continuous integer-order baseline. 
Let $\mathbf{F}_t = \mathbf{F}(\mathbf{x}^l_t)$ denote the latent feature at timestep $t$, and let $\delta$ denote the prediction distance from $t$ to $t-\delta$. 
For a sufficiently smooth feature trajectory, the Taylor expansion of $\mathbf{F}_{t-\delta}$ around $t$ is given by
\begin{equation}
    \mathbf{F}_{t-\delta}
    =
    \sum_{i=0}^{\infty}
    \frac{(-\delta)^i}{i!}
    \frac{\partial^i \mathbf{F}_t}{\partial t^i}.
\end{equation}
By defining the temporal derivative operator as
\begin{equation}
    \nabla_t = \frac{\partial}{\partial t},
\end{equation}
the above expansion can be written in operator form as
\begin{equation}
    \mathbf{F}_{t-\delta}
    =
    \left(
    \sum_{i=0}^{\infty}
    \frac{(-\delta \nabla_t)^i}{i!}
    \right)
    \mathbf{F}_t
    =
    e^{-\delta \nabla_t}\mathbf{F}_t .
\end{equation}
Equivalently, by defining
\begin{equation}
    z = -\delta \nabla_t ,
\end{equation}
the integer-order extrapolation can be interpreted as applying the exponential shift operator:
\begin{equation}
    \mathbf{F}_{t-\delta}
    \approx
    e^z \mathbf{F}_t ,
    \label{eq:appendix_operator_equivalence}
\end{equation}
where the approximation emphasizes the practical setting in which the continuous operator is implemented through finite-order truncation and discrete feature differences. 

\paragraph{Continuous Fractional Generalization.}
To model the anomalous and stage-dependent evolution of DiT features, FADE replaces the classical exponential operator with a Mittag--Leffler (ML) operator. 
The ML function is defined as
\begin{equation}
    E_{\gamma}(z)
    =
    \sum_{i=0}^{\infty}
    \frac{z^i}{\Gamma(\gamma i + 1)},
\end{equation}
where $\gamma$ controls the fractional propagation order. 
Substituting the exponential shift variable $z = -\delta \nabla_t$ into the ML operator gives
\begin{equation}
    E_{\gamma}(-\delta \nabla_t)\mathbf{F}_t
    =
    \sum_{i=0}^{\infty}
    \frac{(-\delta \nabla_t)^i}{\Gamma(\gamma i + 1)}
    \mathbf{F}_t .
\end{equation}
By isolating the zero-order term and using $\Gamma(1)=1$, we obtain
\begin{equation}
    E_{\gamma}(-\delta \nabla_t)\mathbf{F}_t
    =
    \mathbf{F}_t
    +
    \sum_{i=1}^{\infty}
    \frac{(-\delta)^i}{\Gamma(\gamma i + 1)}
    \nabla_t^i \mathbf{F}_t .
\end{equation}

\paragraph{Truncation and Numerical Discretization.}
For tractable inference, the infinite fractional series is truncated to a finite order $m$. 
Meanwhile, since diffusion sampling proceeds over discrete timesteps, the continuous derivative $\nabla_t^i \mathbf{F}_t$ is approximated by the $i$-th order discrete feature difference, denoted by $\Delta^i \mathbf{F}_t$. 
The backward difference operator is recursively defined as
\begin{align}
    \Delta^1 \mathbf{F}_t
    &=
    \mathbf{F}_t - \mathbf{F}_{t-1}, \\
    \Delta^i \mathbf{F}_t
    &=
    \Delta^{i-1} \mathbf{F}_t
    -
    \Delta^{i-1} \mathbf{F}_{t-1},
    \quad i \ge 2 .
\end{align}
Substituting the continuous derivatives with discrete feature differences yields the FADE predictive formulation:
\begin{equation}
    \mathbf{F}_{\mathrm{pred}}(\mathbf{x}^l_{t-\delta})
    =
    \mathbf{F}(\mathbf{x}^l_t)
    +
    \sum_{i=1}^{m}
    \frac{
    \Delta^i \mathbf{F}(\mathbf{x}^l_t)
    }{
    \Gamma(\gamma i + 1)
    }
    (-\delta)^i .
    \label{eq:fade_predictive_form}
\end{equation}
Here, $\delta$ denotes the effective prediction distance, instantiated in practice as the skip interval between the latest computed step and the target prediction step.

\paragraph{Degeneration to the Integer-Order Baseline.}
The classical integer-order extrapolation is recovered as a special case when $\gamma = 1$. 
Using the Gamma function identity
\begin{equation}
    \Gamma(i+1) = i!,
\end{equation}
Eq.~\eqref{eq:fade_predictive_form} becomes
\begin{equation}
    \mathbf{F}_{\mathrm{pred}}(\mathbf{x}^l_{t-\delta})
    \big|_{\gamma=1}
    =
    \mathbf{F}(\mathbf{x}^l_t)
    +
    \sum_{i=1}^{m}
    \frac{
    \Delta^i \mathbf{F}(\mathbf{x}^l_t)
    }{
    i!
    }
    (-\delta)^i .
\end{equation}
Therefore, the conventional integer-order polynomial forecasting rule can be viewed as the $\gamma=1$ instance of the more general FADE formulation, while $\gamma \neq 1$ provides a flexible coefficient law for modeling anomalous, stage-dependent feature dynamics.

\section{Implementation Details of Stage-Aware Calibration}
\label{sec:calibration_details}
\subsection{Model-Specific Calibration and Extreme Data Efficiency}
A significant advantage of FADE is its extreme data efficiency during the offline calibration phase. We construct the calibration dataset $\mathcal{D}_{\text{calib}}$ using only \textbf{10 randomly sampled conditions} paired with standard Gaussian noise initializations. To ensure the calibration perfectly aligns with the native inference pipelines of different generation paradigms, the conditioning inputs are specifically configured as follows:

\begin{itemize}
    \item \textbf{Class-Conditional Models (e.g., DiT-XL/2):} 
    $\mathcal{D}_{\text{calib}}$ is constructed by randomly sampling 10 class labels from the ImageNet training label space, coupled with random initial noise seeds. The calibration seeds are kept disjoint from those used for final evaluation.
    \item \textbf{Text-to-Image Models (e.g., FLUX.1):} 
    The calibration conditions are formed by randomly selecting 10 diverse text prompts from the MS-COCO training split, paired with random initial seeds. These calibration prompts and seeds are not reused in the final DrawBench evaluation.
    \item \textbf{Text-to-Video Models (e.g., HunyuanVideo):} 
For video generation, $\mathcal{D}_{\text{calib}}$ is constructed using 10 randomly sampled prompts from T2V-CompBench, together with random initial noise seeds. These calibration prompts and seeds are kept disjoint from those used in the final VBench evaluation.
\end{itemize}

Despite the limited sample size ($N=10$) and the inherent stochasticity of diverse semantic conditions, the derived Look-Up Table (LUT) for $\hat{\gamma}_{t,s}$ captures a stable profile of latent dynamics that generalizes effectively to unseen prompts during online inference. 

\subsection{Stage-Aware Calibration Procedure}

Calibration follows the online compute--predict schedule using
a closed-loop procedure, as summarized in
Algorithm~\ref{alg:stage_aware_calibration}.
We approximate the calibration objective in
Eq.~\eqref{eq:gamma_optimization} through sample-wise directional
grid searches followed by aggregation.
At each prediction location $(t,s)$, candidate orders in
$\mathcal{G}$ are evaluated using
Eqs.~\eqref{eq:fade_extrapolation} and
\eqref{eq:directional_loss}. The prediction associated with
the minimum directional loss is injected into the residual stream:
\begin{equation}
    \mathbf{H}_{t,s}
    \leftarrow
    \mathbf{H}_{t,s}
    +
    \mathrm{gate}_{t,s}\odot\hat{\mathbf{F}}^*_{t,s},
\end{equation}
where $\mathbf{H}_{t,s}$ is the current hidden residual state
and $\mathrm{gate}_{t,s}$ is the corresponding model gate.
Subsequent calibration therefore accounts for recursively
propagated prediction errors.

For each timestep--location pair $(t,s)$, the selected orders
are collected across calibration samples and aggregated to obtain
$\gamma^*_{t,s}$. The refined order
$\hat{\gamma}_{t,s}=\lambda\gamma^*_{t,s}$ is then stored in the LUT.
The feature cache contains the zeroth-order feature
$\Delta^0\mathbf{F}_{t_c,s}=\mathbf{F}_{t_c,s}$ and its finite
differences, where $t_c$ denotes the latest full-compute timestep.

\begin{algorithm}[t]
\caption{Stage-Aware Closed-Loop Offline Calibration for FADE}
\label{alg:stage_aware_calibration}
\begin{algorithmic}[1]
\STATE \textbf{Input:} calibration dataset
$\mathcal{D}_{\mathrm{calib}}$, candidate grid $\mathcal{G}$,
compute-step schedule $\mathcal{C}$ including warm-up,
maximum order $m$, refinement factor $\lambda$
\STATE \textbf{Output:} refined fractional-order table
$\mathrm{LUT}[t,s]$
\STATE Initialize an empty collection $\mathcal{A}_{t,s}$
for each target pair $(t,s)$

\FOR{each calibration sample
$\mathbf{q}\in\mathcal{D}_{\mathrm{calib}}$}
    \STATE Initialize the diffusion latent and conditioning
    from $\mathbf{q}$

    \FOR{each denoising timestep $t$}
        \IF{$t\in\mathcal{C}$}
            \STATE Run the standard full forward pass to
            obtain model output $\mathbf{y}_t$
            \STATE Update feature histories and caches
            $\mathcal{H}_s
            \gets\{\Delta^i\mathbf{F}_{t,s}\}_{i=0}^{m}$
            for all target locations $s$
            \STATE $t_c\gets t$
        \ELSE
            \STATE Compute the prediction distance
            $\delta$ from $t_c$ to $t$

            \FOR{each target location $s$ in forward
            execution order}
                \STATE Advance normal forward computations
                to location $s$
                \STATE Obtain the reference feature
                $\mathbf{F}_{\mathrm{GT}}(t,s)$
                by full computation
                \STATE Evaluate all $\gamma\in\mathcal{G}$
                using cache $\mathcal{H}_s$, distance $\delta$,
                and Eq.~\eqref{eq:fade_extrapolation}
                \STATE Select $\gamma^*_{t,s}(\mathbf{q})$
                minimizing Eq.~\eqref{eq:directional_loss}
                and retain its prediction
                $\hat{\mathbf{F}}^*_{t,s}$
                \STATE Append $\gamma^*_{t,s}(\mathbf{q})$
                to $\mathcal{A}_{t,s}$
                \STATE $\mathbf{H}_{t,s}
                \gets
                \mathbf{H}_{t,s}
                +
                \mathrm{gate}_{t,s}
                \odot\hat{\mathbf{F}}^*_{t,s}$
            \ENDFOR

            \STATE Complete the remaining forward computations
            to obtain model output $\mathbf{y}_t$
        \ENDIF

        \STATE Advance the diffusion latent using
        $\mathbf{y}_t$ and the native sampler
    \ENDFOR
\ENDFOR

\FOR{each pair $(t,s)$ with
$\mathcal{A}_{t,s}\neq\emptyset$}
    \STATE $\gamma^*_{t,s}
    \gets\mathrm{Aggregate}(\mathcal{A}_{t,s})$
    \STATE $\mathrm{LUT}[t,s]
    \gets\lambda\gamma^*_{t,s}$
\ENDFOR

\STATE \textbf{return} $\mathrm{LUT}$
\end{algorithmic}
\end{algorithm}

\subsection{Dissecting Trajectory Dynamics: Implicit Regularization vs. Dynamic Rectification}
\label{sec:appendix_trajectory_dynamics}

To rigorously validate the efficacy of the optimal fractional orders found during the aforementioned offline calibration, we shift our focus to the long-horizon error propagation during online inference. By injecting the extrapolated features back into the recursive forward pass, we compare our FADE framework against the Ground-Truth full computation (GT-FI) and Standard Taylor extrapolation (ST-FI).

\subsubsection{The Illusion of Implicit Regularization in Taylor Expansion}
A critical observation in our closed-loop experiments is the highly non-linear nature of the diffusion trajectory. Due to the inherent architectural complexity of Diffusion Transformers (e.g., the global receptive field of self-attention mechanisms), the true latent manifold (GT-FI) exhibits dense, high-frequency structural transitions and abrupt numerical spikes across layers. 

However, when employing the standard Taylor expansion (ST-FI) for caching, we observe a phenomenon we term \textbf{Implicit Regularization}. The strict integer-order assumption ($\gamma=1.0$) acts as a rigid mathematical constraint, aggressively smoothing out these necessary trajectory spikes. It forces the feature evolution into a sub-optimal, overly-regularized path. While this may locally suppress extreme outliers, it fundamentally destroys the high-frequency semantics required for complex generation, inevitably leading to catastrophic directional divergence in later stages.

\subsubsection{FADE: Dual-Stage Rectification under Recursive Error}
In contrast, FADE avoids this rigid regularization by treating the fractional order as a dynamic, stage-aware degree of freedom. By explicitly simulating the recursive error accumulation during the offline calibration, FADE introduces a \textbf{Dual-Stage Rectification} mechanism to faithfully reconstruct the GT trajectory:

\begin{itemize}
    \item \textbf{Directional Rectification via Fractional Order:} Instead of suppressing the structural spikes, the optimized Mittag-Leffler parameters ($\gamma^*$) flexibly adapt to the local curvature of the manifold. By strictly monitoring the cosine similarity (Algorithm~\ref{alg:stage_aware_calibration}), FADE actively corrects the \textit{directional drift} that plagues standard caching methods.
    \item \textbf{Intrinsic--Extrinsic Mismatch Correction:} 
The stage-aware LUT provides state-specific fractional orders, but practical inference still uses a proxy distance instead of the intrinsic propagation scale. 
To account for this residual propagation mismatch, FADE applies a global refinement factor $\lambda$ to obtain $\hat{\gamma}_{t,s}=\lambda\gamma^*_{t,s}$, while preserving the state-specific structure encoded in the LUT.
\end{itemize}

Through this decoupled rectification strategy, FADE is able to faithfully trace the abrupt fluctuations of the ground-truth manifold, completely bypassing the implicit regularization trap of Euclidean-based forecasting methods.

\subsection{Empirical Validation of High Sample Consistency on DiT}
To rigorously justify the use of a minuscule calibration set, we conducted an extensive empirical study focusing on the consistency of the optimal fractional order specifically within the \textbf{DiT-XL/2} architecture. We computed the optimal $\gamma^*_{t,s}$ across 10 independent random noise initializations on ImageNet-512. 

As summarized in Table \ref{tab:gamma_variance_attn}, the optimal fractional order at the stable denoising phase ($t=40$) exhibits a clear layer-dependent pattern. While the early and middle layers naturally display moderate variance to adapt to diverse initial noise inputs, the deep layers (Layers 19--27) maintain strong cross-sample consistency. Most importantly, empirical generation results confirm that the moderate standard deviation observed in the shallow stages has \textbf{negligible impact} on the final perceptual quality. The stable deeper layers effectively absorb these initial fluctuations, ensuring that a single offline calibration remains highly robust across any unseen prompts.

\begin{table}[ht]
\centering
\caption{Cross-sample consistency of the optimal fractional order $\gamma^*$ in DiT attention layers. We highlight the statistics at the most stable denoising phase (Timestep $t=40$). The consistent behavior in deep layers and its robustness to early-stage variance formally validate our offline calibration strategy.}
\label{tab:gamma_variance_attn}
\vspace{2mm}
\resizebox{0.75\textwidth}{!}{
\begin{tabular}{l l ccc}
\toprule
\multirow{2}{*}{\textbf{Network Phase}} & \multirow{2}{*}{\textbf{Layer Indices}} & \multicolumn{3}{c}{\textbf{Attention Module (Attn)}} \\
\cmidrule(lr){3-5}
& & \textbf{Mean $\gamma^*$} & \textbf{Std ($\sigma$)} & \textbf{MAD} \\
\midrule
Early Layers    & Layers 0--9     & 0.64 & 0.144 & 0.128 \\
Middle Layers   & Layers 10--18   & 0.96 & 0.129 & 0.119 \\
Deep Layers     & Layers 19--27   & 0.46 & 0.128 & 0.109 \\
\midrule
\rowcolor{gray!10} 
\textbf{Global Summary} & \textbf{All Layers} & \textbf{0.68} & 0.120 & 0.123 \\
\bottomrule
\end{tabular}
}
\end{table}

\section{Error Propagation and Stability Analysis}
\label{sec:error_analysis}

\paragraph{Scope of the analysis.}
We provide a local and conditional error analysis for the regime
where the local latent evolution is better described by an
intrinsic fractional propagation law.
The goal is not to prove a universal advantage of FADE over all
possible predictors, but to explain why a state-adaptive fractional
coefficient law can reduce the mismatch between the intrinsic
feature trajectory and its practical approximation based on an
extrinsic prediction scale.

\paragraph{Notation.}
Let $t$ denote the denoising timestep and $s$ denote the latent
state index.
We write
\[
\mathbf{F}_{t,s}:=\mathbf{F}(\mathbf{x}_t;s),
\]
and let $\Delta^i\mathbf{F}_{t,s}$ denote the $i$-th order backward
finite difference computed from cached historical features at
state $s$.
For compactness, define the fractional coefficient
\begin{equation}
    b_i(\gamma)
    :=
    \frac{1}{\Gamma(\gamma i+1)}.
    \label{eq:bi_definition}
\end{equation}
Here, $\delta$ denotes the extrinsic prediction scale used in
practical inference, while $\delta_{\mathcal{M},t,s}$ denotes
the intrinsic propagation scale along the local feature trajectory.
The difference between these scales gives rise to a potential
intrinsic--extrinsic propagation mismatch.

\paragraph{Assumption 1: intrinsic fractional local evolution.}
For each timestep--state pair $(t,s)$, assume that there exists
a local intrinsic fractional order $\gamma^\dagger_{t,s}>0$
such that the ground-truth local evolution admits the following
truncated intrinsic expansion:
\begin{equation}
\mathbf{F}^{\mathrm{GT}}_{t-\delta,s}
=
\mathbf{F}_{t,s}
+
\sum_{i=1}^{m}
b_i(\gamma^\dagger_{t,s})
\Delta^i \mathbf{F}_{t,s}
(-\delta_{\mathcal{M},t,s})^i
+
\mathbf{R}^{(t,s)}_{m+1},
\label{eq:appendix_intrinsic_expansion}
\end{equation}
where the truncation remainder satisfies
\begin{equation}
    \left\|
    \mathbf{R}^{(t,s)}_{m+1}
    \right\|
    \le
    C_{t,s}\delta_{\mathcal{M},t,s}^{m+1}.
    \label{eq:appendix_remainder_bound}
\end{equation}
This assumption is local and conditional: it only states that
the nearby feature evolution can be approximated by a truncated
intrinsic ML-type expansion.

\paragraph{FADE predictor.}
FADE uses the extrinsic prediction scale $\delta$ and the
globally refined fractional order
\begin{equation}
    \hat{\gamma}_{t,s}
    =
    \lambda\gamma^*_{t,s},
    \label{eq:appendix_refined_order}
\end{equation}
where $\gamma^*_{t,s}$ is obtained from offline stage-aware
calibration.
The corresponding predictor is
\begin{equation}
\widehat{\mathbf{F}}^{\mathrm{FADE}}_{t-\delta,s}
=
\mathbf{F}_{t,s}
+
\sum_{i=1}^{m}
b_i(\lambda\gamma^*_{t,s})
\Delta^i \mathbf{F}_{t,s}
(-\delta)^i.
\label{eq:appendix_fade_predictor}
\end{equation}

\paragraph{Theorem 1: residual-aware FADE error bound.}
Under Assumption 1, the prediction error of FADE satisfies
\begin{align}
\left\|
\mathbf{F}^{\mathrm{GT}}_{t-\delta,s}
-
\widehat{\mathbf{F}}^{\mathrm{FADE}}_{t-\delta,s}
\right\|
&\le
C_{t,s}\delta_{\mathcal{M},t,s}^{m+1}
\nonumber\\
&\quad+
\sum_{i=1}^{m}
\left\|
\Delta^i\mathbf{F}_{t,s}
\right\|
\left|
\frac{(-\delta_{\mathcal{M},t,s})^i}
{\Gamma(\gamma^\dagger_{t,s}i+1)}
-
\frac{(-\delta)^i}
{\Gamma(\lambda\gamma^*_{t,s}i+1)}
\right|.
\label{eq:appendix_fade_error_bound}
\end{align}

\paragraph{Proof.}
Subtracting Eq.~\eqref{eq:appendix_fade_predictor} from
Eq.~\eqref{eq:appendix_intrinsic_expansion} gives
\begin{align}
\mathbf{F}^{\mathrm{GT}}_{t-\delta,s}
-
\widehat{\mathbf{F}}^{\mathrm{FADE}}_{t-\delta,s}
&=
\mathbf{R}^{(t,s)}_{m+1}
\nonumber\\
&\quad+
\sum_{i=1}^{m}
\left[
b_i(\gamma^\dagger_{t,s})(-\delta_{\mathcal{M},t,s})^i
-
b_i(\lambda\gamma^*_{t,s})(-\delta)^i
\right]
\Delta^i\mathbf{F}_{t,s}.
\end{align}
Taking norms and applying the triangle inequality yields
\begin{align}
\left\|
\mathbf{F}^{\mathrm{GT}}_{t-\delta,s}
-
\widehat{\mathbf{F}}^{\mathrm{FADE}}_{t-\delta,s}
\right\|
&\le
\left\|
\mathbf{R}^{(t,s)}_{m+1}
\right\|
\nonumber\\
&\quad+
\sum_{i=1}^{m}
\left\|
\Delta^i\mathbf{F}_{t,s}
\right\|
\left|
b_i(\gamma^\dagger_{t,s})(-\delta_{\mathcal{M},t,s})^i
-
b_i(\lambda\gamma^*_{t,s})(-\delta)^i
\right|.
\end{align}
Using Eq.~\eqref{eq:appendix_remainder_bound} and the definition
of $b_i(\cdot)$ gives Eq.~\eqref{eq:appendix_fade_error_bound}.
\hfill $\square$

\paragraph{Decomposing the mismatch term.}
The second term in Eq.~\eqref{eq:appendix_fade_error_bound}
contains both intrinsic--extrinsic scale mismatch and
coefficient mismatch.
Indeed, for each order $i$, we have
\begin{align}
&
\left|
b_i(\gamma^\dagger_{t,s})(-\delta_{\mathcal{M},t,s})^i
-
b_i(\lambda\gamma^*_{t,s})(-\delta)^i
\right|
\nonumber\\
&\le
\underbrace{
\left|
b_i(\gamma^\dagger_{t,s})
\right|
\left|
\delta_{\mathcal{M},t,s}^{i}
-
\delta^i
\right|
}_{\text{intrinsic--extrinsic scale mismatch}}
+
\underbrace{
\delta^i
\left|
b_i(\gamma^\dagger_{t,s})
-
b_i(\lambda\gamma^*_{t,s})
\right|
}_{\text{fractional coefficient mismatch}}.
\label{eq:appendix_mismatch_decomposition}
\end{align}
This shows explicitly that FADE can reduce the coefficient
mismatch by calibrating $\gamma^*_{t,s}$ and refining it with
$\lambda$, while the remaining scale discrepancy arises from
the difference between the extrinsic prediction scale $\delta$
and the intrinsic propagation scale $\delta_{\mathcal{M},t,s}$.

\paragraph{Smoothness of the coefficient law.}
Assume that the relevant fractional orders lie in a compact interval
\[
I=[\gamma_{\min},\gamma_{\max}]\subset(0,\infty).
\]
Since $b_i(\gamma)=1/\Gamma(\gamma i+1)$ is smooth on $I$,
there exists a finite constant
\begin{equation}
L_i
:=
\sup_{\gamma\in I}
\left|
\frac{d}{d\gamma}b_i(\gamma)
\right|
<\infty,
\label{eq:appendix_lipschitz_constant}
\end{equation}
such that
\begin{equation}
\left|
b_i(\gamma_1)-b_i(\gamma_2)
\right|
\le
L_i|\gamma_1-\gamma_2|,
\qquad
\forall\gamma_1,\gamma_2\in I.
\label{eq:appendix_lipschitz}
\end{equation}
Therefore, Eq.~\eqref{eq:appendix_mismatch_decomposition} implies
\begin{align}
&
\left|
b_i(\gamma^\dagger_{t,s})(-\delta_{\mathcal{M},t,s})^i
-
b_i(\lambda\gamma^*_{t,s})(-\delta)^i
\right|
\nonumber\\
&\le
\left|
b_i(\gamma^\dagger_{t,s})
\right|
\left|
\delta_{\mathcal{M},t,s}^{i}
-
\delta^i
\right|
+
L_i
\left|
\gamma^\dagger_{t,s}
-
\lambda\gamma^*_{t,s}
\right|
\delta^i.
\label{eq:appendix_lipschitz_mismatch}
\end{align}
This bound separates the error into an intrinsic--extrinsic
scale mismatch and an order-induced coefficient mismatch.

\paragraph{Taylor predictor.}
The vanilla Taylor predictor corresponds to the special case
$\gamma=1$, since
\[
b_i(1)
=
\frac{1}{\Gamma(i+1)}
=
\frac{1}{i!}.
\]
Its prediction form is
\begin{equation}
\widehat{\mathbf{F}}^{\mathrm{Taylor}}_{t-\delta,s}
=
\mathbf{F}_{t,s}
+
\sum_{i=1}^{m}
\frac{1}{i!}
\Delta^i \mathbf{F}_{t,s}
(-\delta)^i.
\label{eq:appendix_taylor_predictor}
\end{equation}
Following the same derivation as above, we obtain
\begin{align}
\left\|
\mathbf{F}^{\mathrm{GT}}_{t-\delta,s}
-
\widehat{\mathbf{F}}^{\mathrm{Taylor}}_{t-\delta,s}
\right\|
&\le
C_{t,s}\delta_{\mathcal{M},t,s}^{m+1}
\nonumber\\
&\quad+
\sum_{i=1}^{m}
\left\|
\Delta^i\mathbf{F}_{t,s}
\right\|
\left|
\frac{(-\delta_{\mathcal{M},t,s})^i}
{\Gamma(\gamma^\dagger_{t,s}i+1)}
-
\frac{(-\delta)^i}{i!}
\right|.
\label{eq:appendix_taylor_error_bound}
\end{align}

\paragraph{Conditional advantage over Taylor.}
The truncation term in Eq.~\eqref{eq:appendix_fade_error_bound}
and Eq.~\eqref{eq:appendix_taylor_error_bound} is shared.
Therefore, FADE has a tighter derived upper bound than Taylor
whenever
\begin{align}
\sum_{i=1}^{m}
\left\|
\Delta^i\mathbf{F}_{t,s}
\right\|
\left|
\frac{(-\delta_{\mathcal{M},t,s})^i}
{\Gamma(\gamma^\dagger_{t,s}i+1)}
-
\frac{(-\delta)^i}
{\Gamma(\lambda\gamma^*_{t,s}i+1)}
\right|
<
\sum_{i=1}^{m}
\left\|
\Delta^i\mathbf{F}_{t,s}
\right\|
\left|
\frac{(-\delta_{\mathcal{M},t,s})^i}
{\Gamma(\gamma^\dagger_{t,s}i+1)}
-
\frac{(-\delta)^i}{i!}
\right|.
\label{eq:appendix_conditional_advantage}
\end{align}
This condition states that the refined fractional coefficient
law must approximate the intrinsic local evolution more closely
than the rigid Taylor coefficient law, accounting for the
difference between intrinsic and extrinsic propagation scales.
It is a local and conditional statement, rather than a universal
guarantee over arbitrary latent processes.

\paragraph{Role of the global refinement factor $\lambda$.}
The offline-calibrated fractional orders $\gamma^*_{t,s}$ may
still exhibit a global scale bias in practical inference.
To account for this potential bias, we introduce a global
refinement factor $\lambda$ and use
\[
\hat{\gamma}_{t,s}=\lambda\gamma^*_{t,s}.
\]
This provides a lightweight global adjustment to the calibrated
orders while preserving the state-specific structure captured
by the offline LUT.

\paragraph{Summary of the local error analysis.}
Under the intrinsic fractional local evolution assumption,
the prediction error is bounded by a truncation term and a
mismatch term involving propagation scales and fractional
coefficients. Taylor extrapolation corresponds to the
$\gamma=1$ case of the fractional coefficient family.
FADE calibrates stage-adaptive orders $\gamma^*_{t,s}$ and
applies the refined orders
$\hat{\gamma}_{t,s}=\lambda\gamma^*_{t,s}$ during inference.
When its mismatch term is smaller than that of Taylor,
FADE has a tighter derived upper bound.
These designs provide a more flexible local approximation
of the intrinsic dynamics, consistent with the improved
robustness observed under large prediction distances.

\section{Comprehensive Analysis of Latent Trajectory Dynamics}
\label{sec:dynamic_analysis}

\subsection{Empirical Analysis of Fractional Dynamics on DiT-XL/2}
\label{subsec:dit_analysis}

\begin{figure}[htbp]
    \centering
    \begin{minipage}[b]{0.45\textwidth}
        \centering
        \includegraphics[width=\textwidth]
        {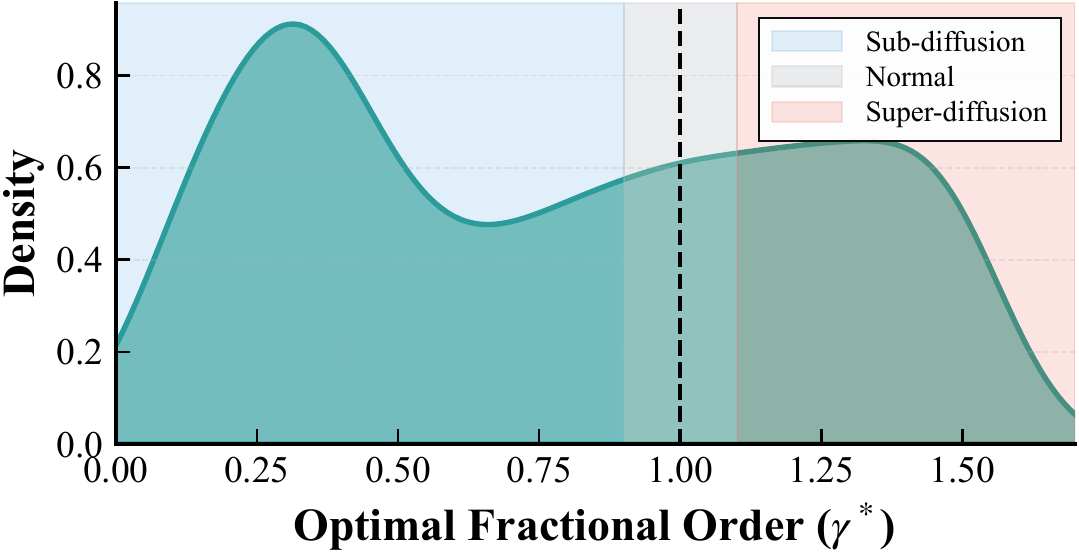}
        \centerline{(a) Global Density Distribution of $\gamma^*$}
    \end{minipage}
    \hfill
    \begin{minipage}[b]{0.52\textwidth}
        \centering
        \includegraphics[width=\textwidth]
        {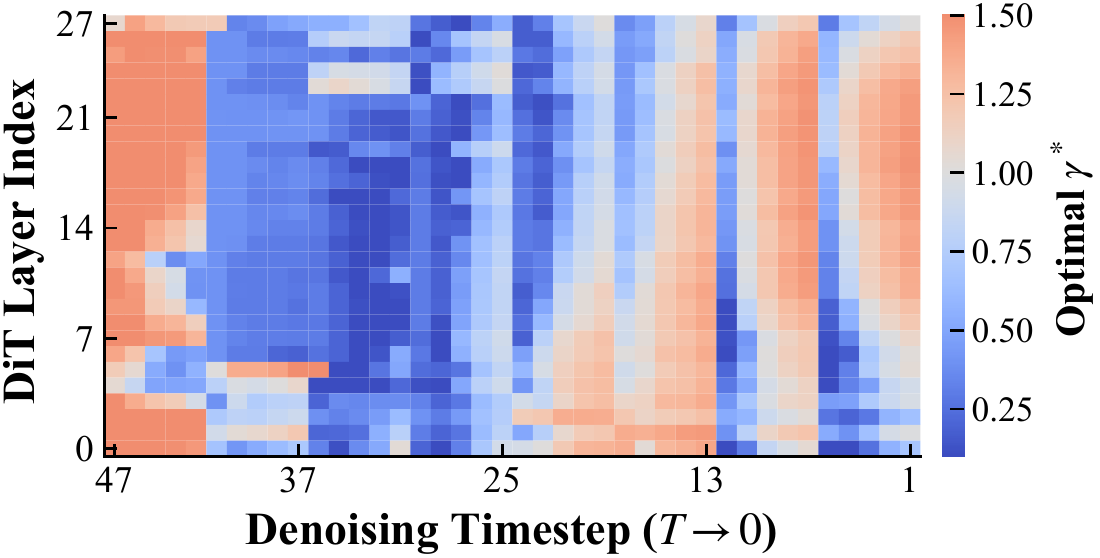}
        \centerline{(b) Spatiotemporal Heatmap of $\gamma^*$}
    \end{minipage}
    \vspace{-0.2cm}
    \caption{\textbf{Spatiotemporal visualization of the optimal
    fractional orders ($\gamma^*$) on DiT-XL/2.}
    (a) Global KDE distribution of $\gamma^*$.
    (b) Variation across denoising timesteps and transformer
    layers, showing a broadly shared U-shaped temporal pattern.}
    \label{fig:appendix_gamma}
\end{figure}

We analyze the optimal fractional orders $\gamma^*$ across
the 28 layers of DiT-XL/2 during a standard 50-step generation
process. Statistics are averaged over 50 randomly sampled
generation trajectories. Figure~\ref{fig:appendix_gamma}
shows the global distribution and spatiotemporal variation
of the calibrated orders.

\paragraph{Global Distribution: Fractional Regimes Dominate.}
As shown in Figure~\ref{fig:appendix_gamma}(a), orders below
$0.9$ occur more frequently than those above $1.1$, while
the near-integer regime ($\gamma\approx1$) accounts for
approximately 13\% of the analyzed states. Offline calibration
therefore frequently selects orders that differ from the
unit order used in standard Taylor extrapolation.

\paragraph{Temporal Heterogeneity: A U-Shaped Phase Pattern.}
Figure~\ref{fig:appendix_gamma}(b) reveals a U-shaped temporal
pattern: lower orders are concentrated in the middle phase,
while higher orders appear more frequently in the early
and late phases.

\paragraph{Spatial Consistency across Network Depth.}
At a given timestep, the calibrated orders remain relatively
aligned across network layers. Together with the temporal
variation, this motivates stage-aware selection of the
forecasting order.

\subsection{Empirical Analysis of Fractional Dynamics on FLUX.1}
\label{subsec:flux_analysis}

\begin{figure}[htbp]
    \centering
    \begin{minipage}[b]{0.48\textwidth}
        \centering
        \includegraphics[width=\textwidth]
        {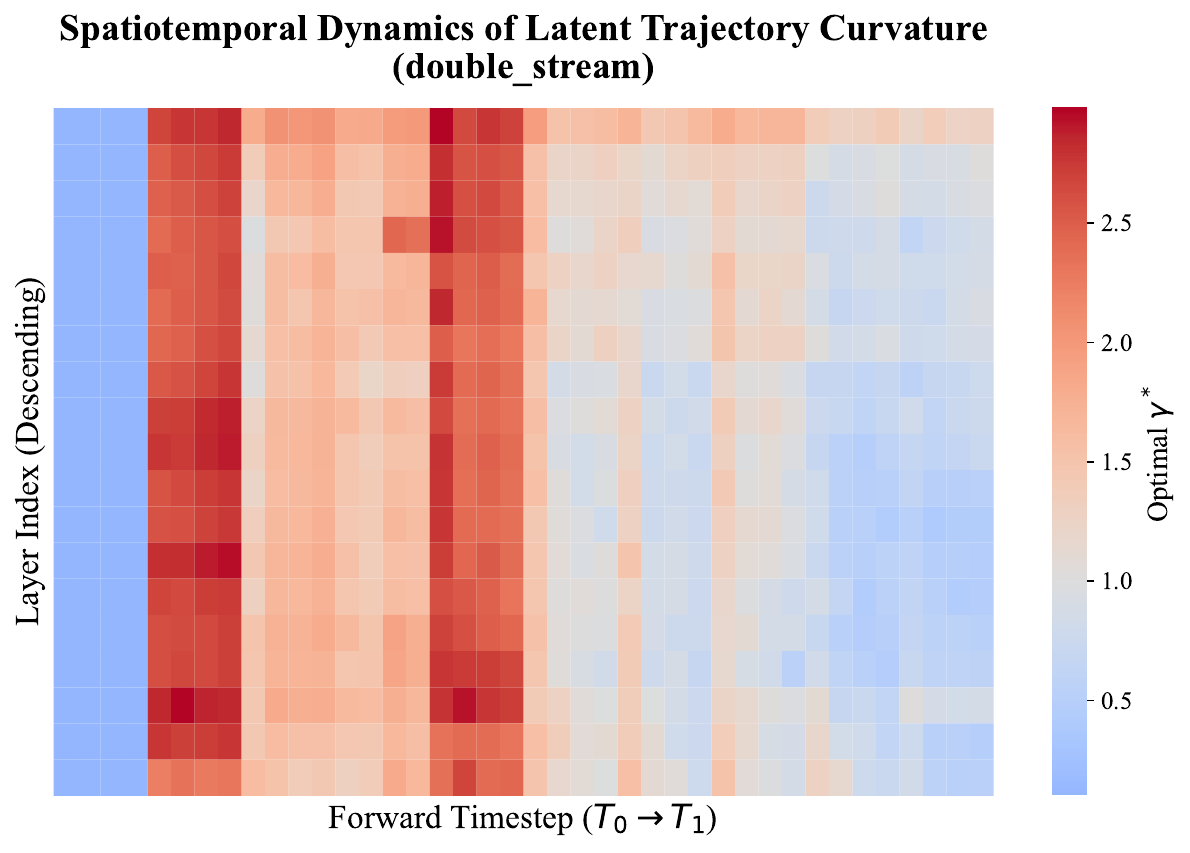}
        \centerline{(a) DoubleStream Blocks}
    \end{minipage}
    \hfill
    \begin{minipage}[b]{0.48\textwidth}
        \centering
        \includegraphics[width=\textwidth]
        {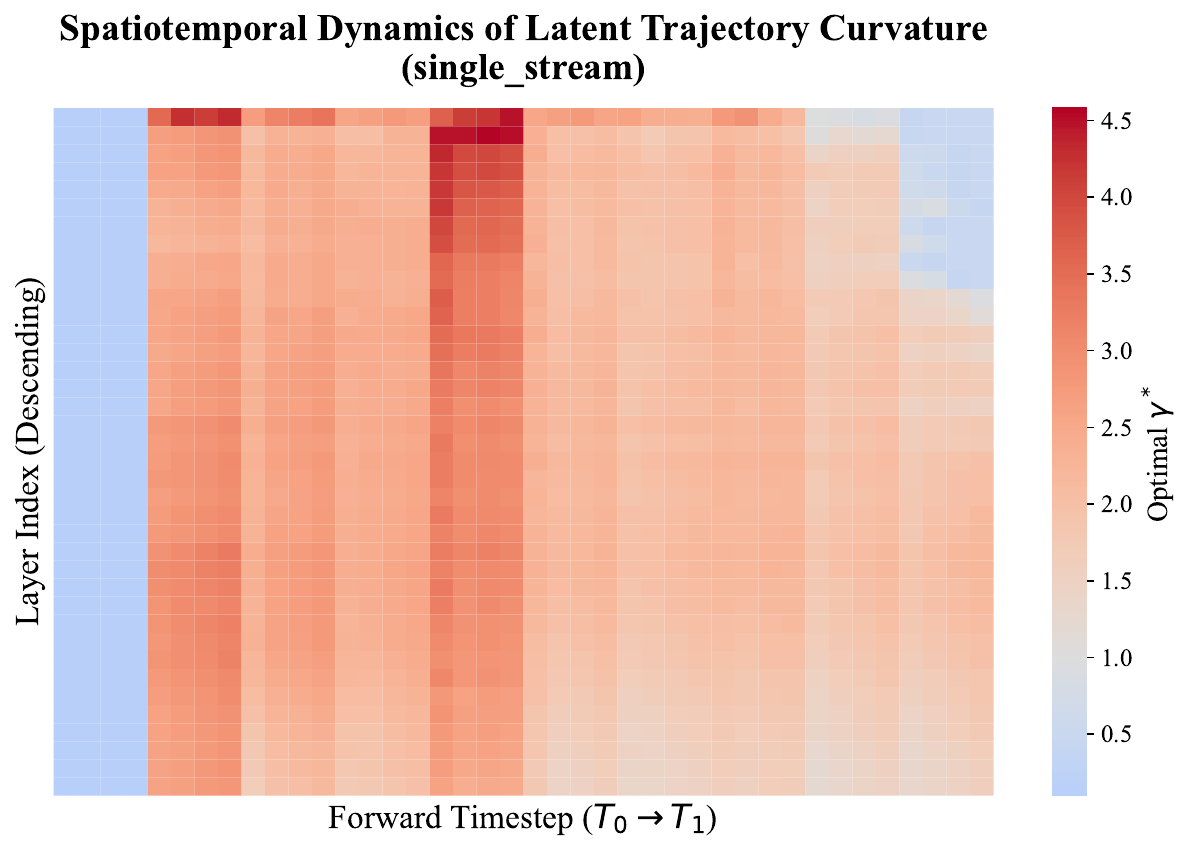}
        \centerline{(b) SingleStream Blocks}
    \end{minipage}
    \vspace{-0.2cm}
    \caption{\textbf{Spatiotemporal decoupling of latent dynamics
    across FLUX.1 architectures.}
    Calibrated orders in (a) DoubleStream and (b) SingleStream
    blocks. SingleStream blocks exhibit a greater proportion
    of higher-order states. Across both streams, higher orders
    are concentrated in the middle stages of $T_0\rightarrow T_1$.}
    \label{fig:flux_heatmap}
\end{figure}

We extend the analysis to FLUX.1, which uses rectified flow
matching and contains distinct \textit{DoubleStream} and
\textit{SingleStream} blocks. For this analysis, we trace
$\gamma^*$ over 50 generation trajectories using text prompts
randomly sampled from DrawBench.
Figure~\ref{fig:flux_heatmap} shows the calibrated orders
across both block types.

\paragraph{Architectural Decoupling: A Shift toward Higher-Order Fractional Regimes.}
In the \textit{DoubleStream} blocks
(Figure~\ref{fig:flux_heatmap}(a)), the higher-order region
($\gamma>1.1$) accounts for 56.8\% of the states, while
orders below $0.9$ account for 32.4\%.
In the \textit{SingleStream} blocks
(Figure~\ref{fig:flux_heatmap}(b)), the higher-order proportion
increases to \textbf{87.0\%}, while the near-integer regime
($\gamma\approx1$) accounts for only approximately 0.5\%.

\paragraph{Temporal Heterogeneity: An Inverted Phase Pattern.}
Along the timeline $T_0\rightarrow T_1$, FLUX exhibits
a temporal pattern opposite to the U-shaped trend observed
in DiT. Relatively lower orders appear in the early and late
phases, while higher orders are concentrated in the middle
phase, as shown in Figure~\ref{fig:flux_heatmap}.

These observations motivate adapting the forecasting order
across both architectural blocks and temporal stages.

\subsection{Empirical Analysis of Fractional Dynamics on HunyuanVideo}
\label{subsec:hunyuan_analysis}

\begin{figure}[htbp]
    \centering
    \begin{minipage}[b]{0.48\textwidth}
        \centering
        \includegraphics[width=\textwidth]
        {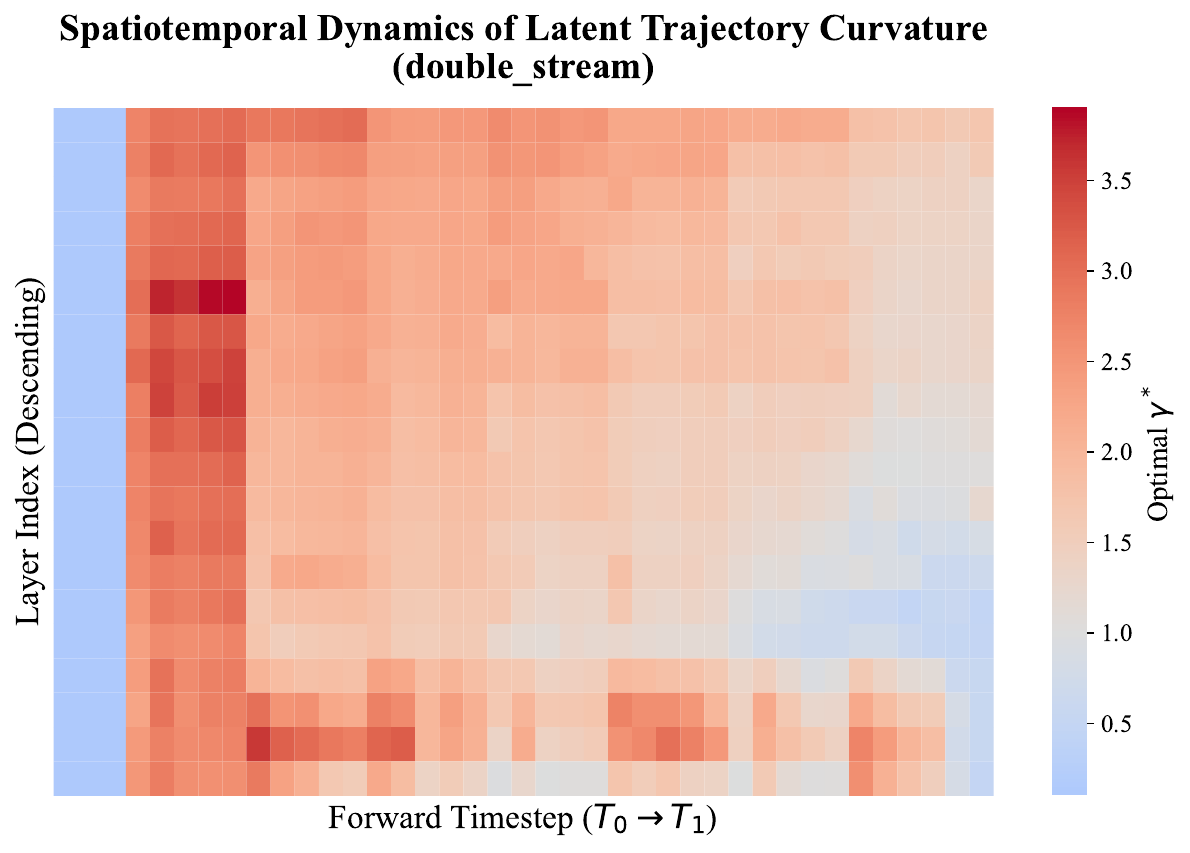}
        \centerline{(a) DoubleStream Blocks}
    \end{minipage}
    \hfill
    \begin{minipage}[b]{0.48\textwidth}
        \centering
        \includegraphics[width=\textwidth]
        {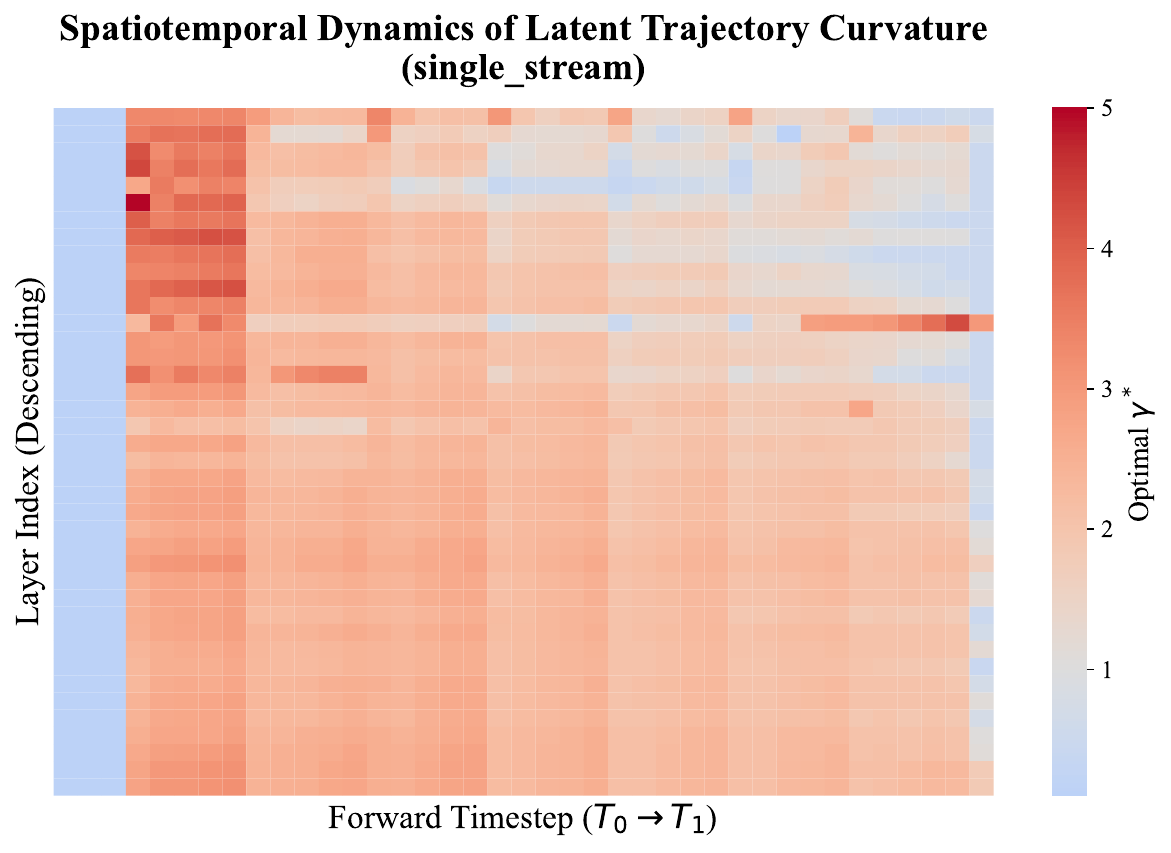}
        \centerline{(b) SingleStream Blocks}
    \end{minipage}
    \vspace{-0.2cm}
    \caption{\textbf{Spatiotemporal decoupling of latent dynamics
    across HunyuanVideo architectures.}
    Both (a) DoubleStream and (b) SingleStream blocks show
    a broad shift toward higher-order regimes ($\gamma>1.1$),
    while near-integer states ($\gamma\approx1$) are
    comparatively uncommon.}
    \label{fig:hunyuan_heatmap}
\end{figure}

We extend the analysis to HunyuanVideo by tracing the optimal
fractional orders $\gamma^*$ across its transformer blocks
using diverse video text prompts.
Figure~\ref{fig:hunyuan_heatmap} presents the distributions
across its DoubleStream and SingleStream blocks.

\paragraph{A Broad Shift toward Super-Diffusion-Like Regimes.}
Figure~\ref{fig:hunyuan_heatmap} shows a broad shift toward
higher-order regimes ($\gamma>1.1$) across both block types.
This trend resembles the behavior observed in the
\textit{SingleStream} blocks of FLUX.1
(Section~\ref{subsec:flux_analysis}).

\paragraph{Limitations of a Uniform Integer-Order Assumption.}
The near-integer regime ($\gamma\approx1$) accounts for only
a small fraction of the calibrated states, while orders below
$0.9$ are also less prominent than in the image DiT analyzed
above. These distributions show that calibration frequently
selects orders beyond the fixed $\gamma=1$ baseline.

Together with the results on DiT and FLUX, these observations
motivate adapting the forecasting order across denoising
stages and model architectures.

\section{Additional Ablation Study}
\label{sec:addition_ablation}

\subsection{Ablation on Fractional-Order Scheduling}
\label{sec:ablation_gamma_schedule}

We further study different fractional-order scheduling strategies on FLUX.1 under the $N=6$ setting. 
This ablation is designed to examine whether a manually specified fixed order or a simple time-dependent fitted schedule can reliably replace stage-aware calibration.

\begin{table}[t]
  \centering
  \caption{Additional ablation on fractional-order scheduling strategies for FLUX.1.}
  \label{tab:flux_gamma_schedule_ablation}
  \setlength{\tabcolsep}{2.6pt}
  \scriptsize
  \resizebox{0.485\linewidth}{!}{
  \begin{tabular}{lccc}
    \toprule
    \textbf{Method} & \textbf{Speed$\uparrow$} & \textbf{IR$\uparrow$} & \textbf{CLIP$\uparrow$} \\
    \midrule
    FADE (fixed $\gamma=1.2$) & 4.99$\times$ & 0.9883 & 27.5061 \\
    FADE (fixed $\gamma=1.6$) & 4.99$\times$ & \textbf{1.0016} & 27.6207 \\
    FADE (fixed $\gamma=1.8$) & 4.99$\times$ & 0.9978 & 27.6239 \\
    FADE (fixed $\gamma=2.0$) & 4.99$\times$ & 0.9936 & 27.6617 \\
    \midrule
    \rowcolor{gray!10}
    FADE (dynamic fit $\gamma(t)$) & 4.99$\times$ & 0.9948 & \textbf{27.7077} \\
    \bottomrule
  \end{tabular}
  }
\end{table}

As shown in Table~\ref{tab:flux_gamma_schedule_ablation}, different fractional-order schedules favor different evaluation metrics. 
A fixed order of $\gamma=1.6$ achieves the best ImageReward, while the dynamic fitted schedule obtains the highest CLIP score. 
This suggests that time-dependent order variation can better align with text-image semantic consistency, but does not uniformly dominate fixed-order schedules across all metrics. 
These results indicate that fractional-order scheduling is sensitive to both the metric and denoising regime, further motivating stage-aware calibration instead of relying on a single manually specified order.

We also evaluate simple fractional-order scheduling strategies on DiT-XL/2. 
This study compares the original integer-order Taylor baseline with a manually tuned fixed fractional order and a fitted time-dependent schedule.

\begin{table}[t]
  \centering
  \caption{Additional ablation on fractional-order scheduling strategies for DiT-XL/2. ``--'' indicates that the metric was not recorded.}
  \label{tab:dit_gamma_schedule_ablation}
  \setlength{\tabcolsep}{3.2pt}
  \scriptsize
  \resizebox{0.52\linewidth}{!}{
  \begin{tabular}{lcc}
    \toprule
    \textbf{Method} & \textbf{FID$\downarrow$} & \textbf{sFID$\downarrow$} \\
    \midrule
    DDIM-50 & 2.29 & 4.29 \\
    HiCache & 2.51 & 5.20 \\
    Taylor & 2.590 & 5.18 \\
    FADE (fixed $\gamma=0.85$) & 2.49 & 4.81 \\
    FADE (Bernoulli-fit $\gamma(t)$) & 2.51 & 5.14 \\
    \bottomrule
  \end{tabular}
  }
\end{table}

As shown in Table~\ref{tab:dit_gamma_schedule_ablation}, both the fixed fractional order and the fitted time-dependent schedule improve over the integer-order Taylor baseline in terms of FID. 
However, the fitted schedule does not consistently outperform the manually tuned fixed order, suggesting that simple parametric schedules can be sensitive to the chosen functional form. 
Together with the FLUX.1 results, this supports the use of stage-aware calibration, which avoids imposing a single fixed or manually fitted $\gamma$ schedule.

\section{Detailed Experimental Protocols}
\label{sec:appendix_experimental_details}

As mentioned in the main text, experiments are conducted on three representative diffusion transformer models across different generative tasks: FLUX.1-dev~\cite{flux2024} for text-to-image generation, HunyuanVideo~\cite{kong2024hunyuanvideo} for text-to-video generation, and DiT-XL/2~\cite{peebles2023scalablediffusionmodelstransformers} for class-conditional image generation. This section provides detailed experimental configurations, including hardware infrastructure, sampling protocols, FADE-specific calibration settings, and baseline hyperparameters.

\subsection{Hardware and Software Infrastructure}

Hardware platforms are selected according to the memory and computation requirements of each model. Specifically, DiT-XL/2 is evaluated on 32GB NVIDIA vGPUs, FLUX.1-dev on NVIDIA H800 GPUs, and HunyuanVideo on NVIDIA A800 GPUs. All models follow their official inference pipelines and default precision settings. Classifier-free guidance (CFG) scales, schedulers, input resolutions, and prompt settings are kept consistent with the standard configurations of the corresponding models.

\subsection{Model Configurations and Sampling Parameters}

Across all evaluated models, official pretrained checkpoints and inference pipelines are strictly followed. Unless otherwise specified, all methods are evaluated under the same number of inference steps, sampling schedules, random seeds, and acceleration ratios for fair comparison.

\begin{itemize}
    \item \textbf{DiT-XL/2 (ImageNet-1K):} The pretrained DiT-XL/2 model is used for class-conditional image generation at a spatial resolution of $256 \times 256$. The inference process is driven by the DDIM sampler with a total of 50 denoising steps.

    \item \textbf{FLUX.1-dev (DrawBench):} The rectified-flow-based FLUX.1-dev model is evaluated for high-fidelity text-to-image generation. Sampling is performed using the default Flow Match Euler scheduler at a resolution of $1024 \times 1024$, with 50 inference steps.

    \item \textbf{HunyuanVideo (VBench):} Text-to-video generation experiments are conducted using the official HunyuanVideo inference pipeline. The default flow-matching sampler is adopted with 50 denoising steps, and all video generation settings follow the official configuration unless otherwise stated.
\end{itemize}

\subsection{FADE and Baseline Configurations}

A detailed configuration scheme for the proposed FADE method and the reproduced baselines is provided below.

\begin{itemize}
    \item \textbf{FADE (Ours):} FADE is implemented as a training-free feature forecasting framework. For each model, we first run an offline calibration stage to profile the optimal fractional order $\gamma^*$ for different denoising stages and network locations. The calibration objective is defined as a directional trajectory matching loss based on cosine distance between the FADE-predicted feature and the corresponding full-computation feature. This choice avoids over-penalizing magnitude variations caused by normalization layers and focuses on the angular consistency of latent feature trajectories.

    The calibrated fractional orders are stored in a static lookup table (LUT), which is indexed by model-specific structural states. For DiT-XL/2, the LUT is indexed by timestep, layer, and module. For FLUX.1-dev, the LUT additionally considers the stream and block structure of the model. For HunyuanVideo, we adopt a coarser LUT indexing strategy due to the substantially higher computational cost and inherent spatiotemporal redundancy of video generation. The LUT is therefore constructed based on timestep and coarse video-transformer block indices. During online inference, FADE directly retrieves the corresponding $\gamma$ from the LUT and performs fractional-order feature extrapolation without any additional optimization or adaptive search.

    \item \textbf{Baselines:} We compare FADE with representative training-free diffusion acceleration methods, including FORA~\cite{selvaraju2024FORAfastforwardcachingdiffusion}, HiCache~\cite{feng2026hicachepluginscaledhermiteupgrade}, and TaylorSeer~\cite{liu2025reusingforecastingacceleratingdiffusion}. All reproduced baselines strictly follow the implementation protocols and default hyperparameter settings reported in their original papers and open-source repositories. For a fair comparison, all baselines are evaluated using the same base model, sampler, number of inference steps, resolution, CFG scale, prompt set, and acceleration schedule as FADE.
\end{itemize}

\paragraph{Offline calibration protocol.}
For each calibration trajectory, a full-computation reference
pass collects the oracle features $\mathbf{F}^{\mathrm{GT}}_{t,s}$
at prediction locations. Given the cached feature basis
$\{\mathbf{D}^{k}_{t,s}\}_{k=0}^{K-1}$ and prediction distance $x$,
each candidate order $\gamma$ is evaluated by
\begin{equation}
\widehat{\mathbf{F}}_{t,s}(\gamma)
=
\sum_{k=0}^{K-1}
\frac{1}{\Gamma(\gamma k+1)}
\mathbf{D}^{k}_{t,s}
x^k,
\label{eq:offline_calibration_prediction}
\end{equation}
where $K$ is the number of retained basis terms and the
zeroth-order coefficient is $1$. The candidate minimizing
the directional error is selected:
\begin{equation}
\gamma^*_{t,s}
=
\arg\min_{\gamma\in\mathcal{G}}
\left[
1-
\cos\left(
\widehat{\mathbf{F}}_{t,s}(\gamma),
\mathbf{F}^{\mathrm{GT}}_{t,s}
\right)
\right].
\label{eq:offline_calibration_objective}
\end{equation}
The selected predictions are propagated using the closed-loop
procedure in Algorithm~\ref{alg:stage_aware_calibration}.

For DiT-XL/2, $s$ indexes the layer and module, and we use
$\mathcal{G}=\{0.1,0.2,\ldots,3.0\}$ with all available
cached orders. For FLUX.1-dev, $s$ additionally specifies
the stream and block, and we use
$\mathcal{G}=\{0.1,0.2,\ldots,5.0\}$ with a two-term cached
basis for numerical stability. HunyuanVideo follows the
FLUX.1-dev protocol with a coarser search and LUT indexed
by timestep and video-transformer block. The calibrated
orders are stored in a static LUT for inference.

\paragraph{Online inference settings.}
For each model and cache interval $N$, we tune $\lambda$
offline and compute the refined orders:
\begin{equation}
\hat{\gamma}_{t,s}
=
\lambda\gamma^*_{t,s}.
\label{eq:online_refined_gamma}
\end{equation}
The refined orders are stored in the LUT and retrieved directly
during inference. We keep $\lambda$ fixed across all evaluation
prompts. The settings are summarized in
Table~\ref{tab:online_fade_settings}.

\begin{table}[h]
\centering
\caption{Online inference settings of FADE. The refinement
factor $\lambda$ is fixed after offline calibration and
introduces no online search overhead.}
\label{tab:online_fade_settings}
\begin{tabular}{lcc}
\toprule
Model & Cache interval $N$ & Refinement factor $\lambda$ \\
\midrule
DiT-XL/2 & 4 & 0.6 \\
DiT-XL/2 & 5 & 0.6 \\
DiT-XL/2 & 6 & 0.7 \\
\midrule
FLUX.1-dev & 5 & 1.3 \\
FLUX.1-dev & 6 & 1.0 \\
FLUX.1-dev & 7 & 1.1 \\
\midrule
HunyuanVideo & default & 1.0 \\
\bottomrule
\end{tabular}
\end{table}

\end{document}